\pdfoutput=1

\documentclass[11pt]{article}

\usepackage[final]{acl}
\usepackage{booktabs}
\usepackage{times}
\usepackage{latexsym}
\usepackage{amsmath} 
\usepackage{longtable}
\usepackage{amsmath}    
\usepackage{amssymb}  
\usepackage{arydshln}
\usepackage{booktabs}

\usepackage[most]{tcolorbox}
\usepackage{xparse}

\usepackage{subcaption}
\usepackage[T1]{fontenc}
\usepackage[ruled,vlined]{algorithm2e}
\usepackage{colortbl}
\usepackage{hyperref} 

\usepackage[utf8]{inputenc}
\usepackage{lipsum}

\usepackage{microtype}
\usepackage{multirow}
\usepackage{graphicx}
\usepackage{amsmath}
\usepackage{inconsolata}

\usepackage{graphicx}
\usepackage{array}
\usepackage{booktabs}
\usepackage{bbding}
\usepackage{pifont}
\usepackage{wasysym}
\usepackage{amssymb}
\usepackage{makecell}
\usepackage[table,dvipsnames]{xcolor} 

\usepackage{amssymb}
\usepackage{pifont}
\usepackage{float}

\newtcolorbox{prompt}[2][]{
    colback=gray!20,
    colframe=white,
    fonttitle=\bfseries\small,
    boxrule=0.4mm,
    fontupper=\small, 
    fontlower=\small,
    coltitle=white,
    title=#2,
    #1,breakable
}
\usepackage{etoolbox}
\usepackage{pgf} 
\definecolor{high}{RGB}{0, 102, 202} 
\definecolor{low}{HTML}{FFFFFF}  
\newcommand*{\opacity}{50}
\newcommand*{\minval}{0.0}
\newcommand*{\maxval}{100}
\newcommand{\gradient}[1]{
    \ifdimcomp{#1pt}{>}{\maxval pt}{#1}{
        \ifdimcomp{#1pt}{<}{\minval pt}{#1}{
            \pgfmathparse{int(round(100*(#1/(\maxval-\minval))-(\minval*(100/(\maxval-\minval)))))}
            \xdef\tempa{\pgfmathresult}
            \cellcolor{high!\tempa!low!\opacity} #1
    }}
}

\newlength{\arrayrulewidthOriginal}

\definecolor{warningcolor}{RGB}{250,35,64}
\definecolor{myLightBlue}{RGB}{200,230,255}
\definecolor{myOrange}{RGB}{255, 116, 23}
\definecolor{myBlue}{RGB}{0, 102, 202}
\definecolor{myGreen}{RGB}{0, 153, 102}   
\definecolor{myRed}{RGB}{204, 0, 0}       
\newcommand\graphic{\raisebox{-2.5pt}{\includegraphics[width=1.0em]{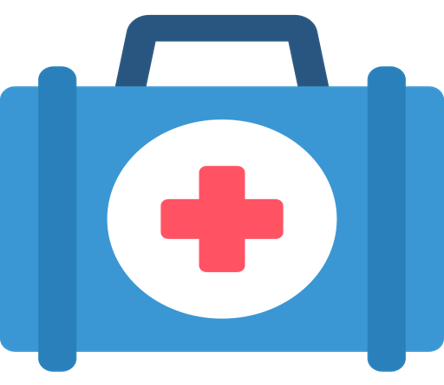}\space}}
\newcommand\pace{\textsc{Pace} }
\newcommand\pacer{\textsc{Pacer} }
\newcommand\pacersft{\textsc{Pacer\textsubscript{SFT}} }
\newcommand\pacerdpo{\textsc{Pacer\textsubscript{SFT+DPO}} }

\title{\graphic PanicToCalm: A Proactive Counseling Agent for Panic Attacks}

\author{Jihyun Lee \textsuperscript{$\clubsuit$},
Yejin Min \textsuperscript{$\clubsuit$},
San Kim \textsuperscript{$\clubsuit$},
Yejin Jeon \textsuperscript{$\clubsuit$},
SungJun Yang\textsuperscript{$\diamondsuit$},
\\
\textbf{ 
Hyounghun Kim\textsuperscript{$\clubsuit,\diamondsuit$}, 
Gary Geunbae Lee\textsuperscript{$\clubsuit,\diamondsuit$}} \\
\\
  \textsuperscript{$\clubsuit$}Graduate School of Artificial Intelligence, POSTECH\\
   \textsuperscript{$\diamondsuit$}Department of Computer Science and Engineering, POSTECH\\
  \small\texttt{\{jihyunlee, yeajinmin, sankm, jeonyj0612, sjyang114, h.kim, gblee\}@postech.ac.kr} \\
}
\begin{document}
\maketitle
\begin{abstract}
Panic attacks are acute episodes of fear and distress, in which timely, appropriate intervention can significantly help individuals regain stability. However, suitable datasets for training such models remain scarce due to ethical and logistical issues. To address this, we introduce \textsc{Pace}, which is a dataset that includes high-distress episodes constructed from first-person narratives, and structured around the principles of Psychological First Aid (PFA). Using this data, we train \textsc{Pacer}, a counseling model designed to provide both empathetic and directive support, which is optimized through supervised learning and simulated preference alignment. To assess its effectiveness, we propose \textsc{PanicEval}, a multi-dimensional framework covering general counseling quality and crisis-specific strategies. Experimental results show that \textsc{Pacer} outperforms strong baselines in both counselor-side metrics and client affect improvement. Human evaluations further confirm its practical value, with \textsc{Pacer} consistently preferred over general, CBT-based, and GPT-4-powered models in panic scenarios \footnote{Code is available at \url{https://github.com/JihyunLee1/PanicToCalm}}.

\end{abstract}


\section{Introduction}
Panic attacks are sudden episodes of intense fear accompanied by physical symptoms such as rapid heartbeat, shortness of breath, and dizziness. These episodes often occur without warning, leaving individuals in acute distress and unable to self-regulate \cite{panic2, apadsm5, panic1}. Immediate intervention from a trained professional through grounding, box-breathing, or relocation to a safe environment, can be highly effective \cite{chalmers2022redevelopment, panic-sol1}.
However, professional support is not always accessible due to time, cost, and availability constraints, which increases the demand for AI systems that are capable of timely assistance. Despite this need, the unpredictable and deeply personal nature of panic attacks makes it nearly impossible to collect authentic, in-the-moment counseling interactions for model training.


\begin{figure}[t]
    \centering
    \includegraphics[width=0.95\columnwidth]{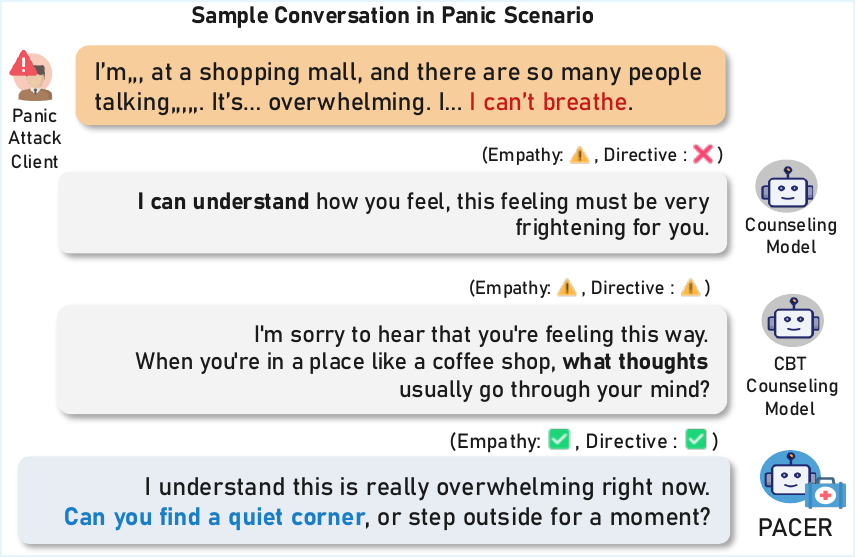}
    \vspace{-5pt}
    \caption{Conversation between a client and different counseling models. \textsc{Pacer} delivers both appropriate empathy and clear, directive guidance to support immediate stabilization.}
    \label{fig:intro}
    \vspace{-10pt}
\end{figure}
Meanwhile, synthetic dialogue datasets with large language models (LLMs) have emerged as a promising alternative in environments where data collection is challenging, given their strong conversational abilities~\cite{dialogcc, soda, synthdst, stark, mixed, picpersona}. In the counseling domain, identifying authentic and diverse sources for dialogue construction such as crowd sourcing~\cite{cbtLLM, smile}, web Q\&A~\cite{healme}, and augmenting from real counseling dialogues~\cite{psych8k}, and applying suitable therapeutic theory~\cite{cbtLLM, cactus, mirror} are key challenges in building effective synthetic datasets. However, prior efforts have predominantly focused on general psychological concerns and employed cognitively demanding counseling procedures, assuming that clients are in relatively stable emotional states. This assumption limits the applicability to high-distress situations like panic attacks, where clients require immediate, directive, and low-cognitive-burden support.

To address this, we introduce \textsc{Pace} (Panic-to-Aid Counseling with Emergency) dataset, that is focused on immediate intervention during acute panic attacks (Figure~\ref{fig:intro}). We begin by collecting first-person narratives from individuals who describe their panic episodes. From these narratives, we identify key components of a panic attack, including environmental context, physical symptoms, emotional responses, and cognitive patterns. As the underlying therapeutic framework, we adopt Psychological First Aid (PFA; ~\citet{brymer}), which is designed to provide immediate stabilization for individuals experiencing high-distress situations such as trauma, PTSD, or panic attacks.

Building on our \textsc{Pace} dataset, we train \textsc{Pacer} (Panic-to-Aid Counseling with Emergency Response), which is a counseling model that is  designed to stabilize acute panic episodes through rapid interventions. To further enhance the model’s ability to cope with a broad spectrum of panic cases, we run simulations using a client agent, exposing it to unpredictable and diverse scenarios. From these simulations, we collect preference feedback from the client agent and perform alignment training to guide the model's behavior toward greater directiveness and empathy.

Finally, we introduce \textsc{PanicEval}, a novel evaluation framework designed to assess counseling models in the context of acute panic interventions. While existing evaluation methods primarily focus on general therapeutic quality, they often fail to capture the urgency and stabilization demands inherent to panic attack scenarios ~\cite{aarons, therapeutic, cactus}. To address this gap, \textsc{PanicEval} incorporates core principles from PFA, which enables the evaluation of both general counseling competencies and first-aid-oriented intervention effectiveness.

During evaluation, we examine two aspects: the counselor’s effectiveness and the client’s emotional improvement. \textsc{Pacer} demonstrates strong performance in both general and first-aid-specific skills, particularly in the clarity of its responses. Significant emotional improvement can also be seen in clients through increased self-esteem and reduced fear. In human evaluations involving both clinical experts and individuals with lived experience, \textsc{Pacer} was consistently preferred over the general, Cognitive Behavioral Therapy (CBT) based, and even closed models like GPT-4o \cite{gpt4o}.

\section{Related Work}
\subsection{Data Challenges in Mental Health}
Building counseling systems requires high-quality dialogue data, but privacy and ethical concerns hinder real-world collection. As such, recent work use closed-source LLMs: HealMe~\cite{healme} generates short CBT-based dialogues from crowdsourced profiles; Cactus~\cite{cactus} adds scripted planning for longer sessions; CPsyCoun~\cite{cpsycoun} reconstructs profiles from anonymized transcripts; and Psych8k~\cite{psych8k}, SMILEChat~\cite{smile}, and CBT-LLM~\cite{cbtLLM} utilize online mental health forums. Yet, these approaches target generalized issues with calm, reflective users. In contrast, we focus on the acute phase of panic attacks, simulating real-time stabilization grounded in PFA.

\subsection{Synthetic Preference Pairs for Value Alignment}
Human-annotated preference datasets have been key to aligning LLMs for better performance and safety \cite{rlhf-hh,learning_to_summarize,webgpt}, but they are costly and hard to scale. To overcome this, recent studies use synthetic preference data generated by LLMs. Methods like R-DPO~\cite{rdpo} and Selfee~\cite{selfee} apply self-critique or model-based scoring, while AutoPM~\cite{autopm} and SAFER-Instruct~\cite{safer_instruct} guide models toward helpfulness, honesty, and harmlessness (HHH) using GPT-4~\cite{gpt4}. Building on these, we apply model-based scoring in a novel setting; outputs are aligned with therapeutic values like empathy and directiveness via interaction with a simulated patient in panic counseling.

\section{Psychological First Aid} 
PFA is a modular intervention developed by the National Child Traumatic Stress Network and the National Center for PTSD to provide support following traumatic events~\cite{brymer}. It is organized around three sequential actions—Look, Listen, and Link—which guides responders in assessing the situation and ensuring safety (Look), stabilizes individuals and offers emotional support (Listen), and connects them to further services and resources (Link) \cite{who}. Unlike long-term therapies like CBT or Emotion-Focused Therapy (EFT; \citet{EFT}), which require reflective engagement over multiple sessions, PFA is designed for acute psychological crises as it prioritizes directive, action-oriented strategies and immediate stabilization. Its effectiveness in managing panic episodes has been highlighted in prior studies \cite{chalmers2022redevelopment}.




\section{\graphic\textsc{Pace}: Panic Episode Dataset with PFA Counseling}

\begin{figure*}[ht]
  \centering
  \includegraphics[width=0.99\linewidth]{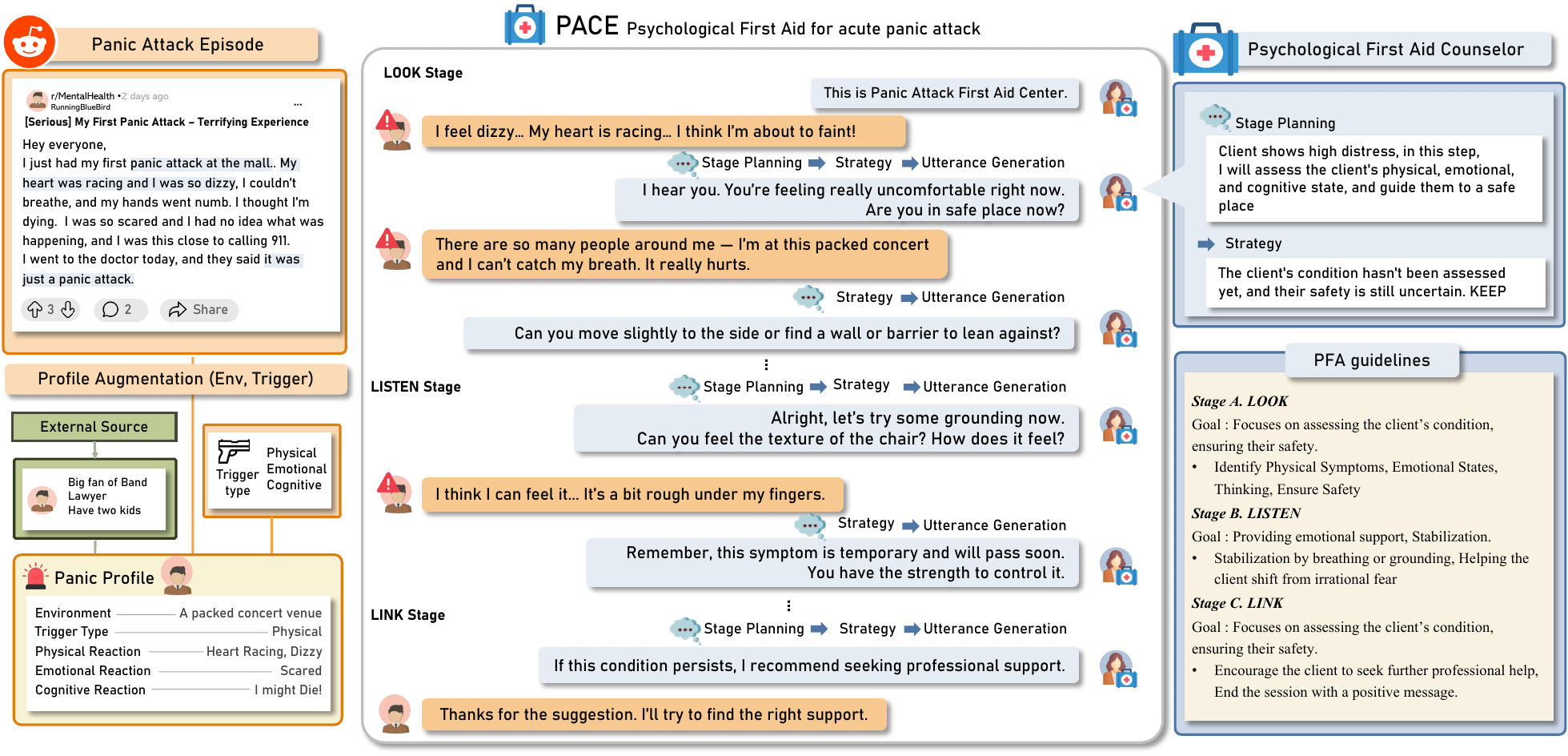}
  \vspace{-7pt}
   \caption{An overview of the automatic pipeline for generating the \pace dataset.}
   \label{fig:main}
   \vspace{-10pt}
\end{figure*}

In this section, we introduce \textsc{Pace}, the first counseling dataset that is designed to address acute panic episodes. We hypothesize that a high-quality dataset should satisfy three key criteria: (1) the client’s episode must be plausible and reflective of real-world panic scenarios; (2) the counselor must successfully guide the client toward stabilization; and (3) the dialogue must maintain general therapeutic standards. 

To meet these requirements, our data generation pipeline consists of three stages: (1) collecting authentic panic narratives, (2) generating multi-staged dialogue scripts based on the PFA guideline, and (3) applying meticulous filtering to ensure counseling quality. We use \textsc{GPT-4o-mini} \cite{gpt4omini} as the primary generation model. An overview is provided in Figure \ref{fig:main}, and corresponding details and prompts are provided in Appendix~\ref{app:pace}, ~\ref{app:pace_example} and ~\ref{app:prompt_for_pace}.

\subsection{Collecting Client Panic Profiles}
As the first step, we model the \textit{panic episode profile}, which is a structured representation of a panic episode consisting of three key components: the contextual Environment, Trigger Type, and Vicious Cycle that is defined as the interlinked physical, emotional, and cognitive responses that reinforce one another, thereby intensifying the severity of the episode~\cite{clark1986cognitive}. The Trigger Type represents the initial stimulus that activates this self-reinforcing cycle. To construct these profiles, we sourced first-person narratives of panic episodes from publicly available social media post and prompted an LLM to extract the relevant components. Given the sensitivity of the data, we applied a meticulous personally identifiable information (PII) removal process using automated detection and manual review (Appendix~\ref{app:pace_crowd}), resulting in 1,226 verified profiles.

To enhance the diversity of contextual settings in the panic profiles, we augmented the Environment and Trigger Type components using an external persona dataset \cite{personalizing}. Specifically, we randomly selected a Trigger Type (physical, emotional, or cognitive) and a persona from the external persona dataset, then prompted an LLM to generate a plausible environment conditioned on these two factors. The generated {Environment} and new Trigger Type were then attached to the original Vicious Cycle to form an augmented profile. For example, given a Trigger Type (e.g., physical) and a persona (e.g., a big fan of live music), the LLM generated a realistic Environment such as “a tightly packed concert venue.” (further examples in Appendix~\ref{app:pace_aug}). Through this process, we constructed a total of 9,997 panic profiles.

\subsection{Defining Counselor Goals}
On the counselor side, we define the model’s goals based on the principles of PFA. Since the original PFA guidelines include actions that are not feasible for dialogue models (e.g., providing physical resources like water), we restructure the goals to focus on interventions that can be delivered through conversation. Specifically, we define the goals for each PFA stage as follows: \textbf{LOOK} identifies the client's panic profile and guides them to a safe environment; \textbf{LISTEN} focuses on stabilization through breathing, grounding and reassurance; and \textbf{LINK} encourages follow-up with a human counselor, ending with on a positive note. The goals for each stage are summarized in Figure~\ref{fig:main} (right) and Appendix~\ref{app:pacer_pfa_goals}, and are denoted as \texttt{goal\textsubscript{stage}}.

\subsection{Structured Dialogue Generation with PFA}

Given a panic profile and the \texttt{goal\textsubscript{stage}}, we generate a complete counseling session that consists of three sequential stages. Our session generation process is guided by two key design choices to ensure alignment with each stage’s therapeutic objective.

First, each stage is generated independently and then concatenated to form the full session. This modular structure facilitates alignment with the objective of each stage. For each stage, we first generate a \texttt{plan} to outline the dialogue flow, and then generate the dialogue turns corresponding to \texttt{plan} and \texttt{goal\textsubscript{stage}}. Second, rather than simulating turn-by-turn interactions between client and counselor agents, we adopt a \textit{script-style} generation approach. In this setup, the model generates the entire dialogue for a stage in a single inference pass, enabling more coherent and goal-aligned outputs within each therapeutic phase. Each stage is represented as: 
\[ 
\texttt{Stage} = \{\texttt{plan}, [(  st_0, s_0, u_0), \dots, ( st_k, s_k, u_k)]\}
\]
where $u_i$ is the client utterance, $s_i$ is the counselor's response, and $st_i$ is the counselor's strategy for generating $s_i$. Strategy $st_i$ reflects the decision of whether the counselor should move on to the next stage or not, based on the current dialogue history and \texttt{goal\textsubscript{stage}}. If $st_i = \texttt{keep}$, the model generates $s_i$, and corresponding $u_i$ continues within the same stage. If $st_i = \texttt{next}$, both $s_i$ and $u_i$ are left empty, and the session transitions to the next stage.



Overall, the LOOK phase is generated given the panic profile and a \texttt{goal\textsubscript{stage}}: $
\texttt{LOOK} = \texttt{LLM}(\text{Profile}, \texttt{goal\textsubscript{stage})}.
$ The subsequent LISTEN and LINK stages are generated based on the previous stage dialogues, profile, and their respective goals: $
\texttt{LISTEN}, \texttt{LINK} = \texttt{LLM}(\text{History}, \text{Profile}, \texttt{goal\textsubscript{stage})}.
$ The final dialogue is formed by concatenating the outputs of the three phases: $
\texttt{Dialogue} = \texttt{LOOK} \oplus \texttt{LISTEN} \oplus \texttt{LINK}
$.


\subsection{Dialogue Quality Filtering} 
\label{sec:pace_filtering}
To ensure structural consistency and counseling quality of the \textsc{Pace} dataset, we applied a multi-step filtering process. First, we removed dialogues with abnormal formatting or excessively long utterances (over 100 words), eliminating 5.92\% of the data. Second, CTRS ~\cite{aarons} is used to assess alignment with general therapeutic standards; five relevant dimensions of Empathy, Clarity, Emotional Alignment, Directive Support, and Encouragement, were rated on a scale from 1 (poor) to 5 (excellent) by prompting GPT-4o. Dialogues that received a score equal to or less than 3 for any dimension were removed (2.42\%), ultimately resulting in 8,263 dialogues and 106,433 turns. Additional validation of CTRS filtering with human agreement is provided in Appendix~\ref{app:pace_ctrs_filtering_human}.

\subsection{Expert Review of \graphic\textsc{Pace}}
\begin{table}[h!]
\centering
\resizebox{\columnwidth}{!}{
\begin{tabular}{lccccc}
\toprule
\textbf{Category} & \textbf{Very Poor} & \textbf{Poor} & \textbf{Fair} & \textbf{Good} & \textbf{Excellent} \\
\midrule
Empathy         & 0.0\% & 1.0\% & 18.0\% & 57.0\% & 24.0\% \\
Clarity         & 0.0\% & 1.0\% & 29.0\% & 50.7\% & 19.3\% \\
Emotional App.  & 0.0\% & 2.7\% & 34.3\% & 48.7\% & 14.3\% \\
Directive       & 0.0\% & 0.0\% & 9.7\%  & 64.0\% & 26.3\% \\
Encouragement   & 0.0\% & 0.0\% & 19.7\% & 56.0\% & 24.3\% \\
\bottomrule
\end{tabular}
}
\caption{Expert evaluation of generated dialogues using the CTRS rubric. 
Scores range from 1 (very poor) to 5 (excellent).}
\label{tab:expert_ctrs}
\end{table}
We conducted an expert evaluation with three mental health professionals, to validate the clinical appropriateness and quality of the generated dialogues 
We randomly sampled 100 dialogues from the training set and asked the experts 
to rate each response using the CTRS rubric on a 5-point scale 
(from 1 = very poor to 5 = excellent). 
The results (Table~\ref{tab:expert_ctrs}) show that most dialogues fall within the fair to excellent range, 
supporting the overall validity of the generated data. 
Specifically, the average scores were 4.04 for Empathy, 
3.88 for Clarity, 3.75 for Emotional Appropriateness, 
4.17 for Directive Support, and 4.05 for Encouragement.

\subsection{Comparison \graphic\textsc{Pace} with Existing Counseling Datasets}

\begin{table*}[t!]
\centering
\resizebox{\textwidth}{!}{
\begin{tabular}{@{}lccccccc@{}}
\toprule
\textbf{Dataset} & \textbf{Counseling Theory} & \textbf{Profile Source} & \textbf{Issue Type} & \textbf{In-Crisis} & \textbf{Language} & \textbf{\# of Dialogues} & \textbf{Multi Turn} \\
\midrule
PsyCon~\cite{psycon}   & Not Specified      & Records,  Crowed Sourcing & 7 psychological issues        & No   & English & 1,020   & 24.58 \\
Psych8k~\cite{psych8k}   & Not Specified      & Counseling Records & Anx, Dep, Relation, Minority        & No   & English & 8,187   & 1.00 \\
SmileChat~\cite{smile}  & Not Specified      & Social Media       & Not Specified        & No   & Chinese & 55,165  & 10.4 \\
HealMe~\cite{healme}   & Cognitive Behavioral Therapy          & Crowed Sourcing    & Not Specified        & No   & English & 1,300   & 3.0 \\
CBT-LLM ~\cite{cbtLLM}            & Cognitive Behavioral Therapy       & Social Media       & Not Specified        & No   & Chinese & 22,327  & 1.0 \\
CACTUS~\cite{cactus}                        & Cognitive Behavioral Therapy          & Crowed Sourcing    & Not Specified        & No   & English & 31,577  & 16.6 \\
\graphic\textsc{Pace}                 & Psychological First Aid & Social Media       & \textbf{Panic Disorder} & \textbf{Yes} & English & 8,263   & 12.8 \\
\bottomrule
\end{tabular}
}
\vspace{-5pt}
\caption{Comparison of counseling dialogue datasets. \textit{Anx} and \textit{Dep} denote Anxiety and Depression, respectively. }
\label{tab:dataset_comparison}
\vspace{-10pt}
\end{table*}


In Table~\ref{tab:dataset_comparison}, we compare our dataset \textsc{Pace} with other existing synthetic counseling datasets. While prior datasets primarily address general mental health concerns or do not specify their target issues, \pace specifically focuses on acute panic attack scenarios. Notably, it is the only dataset that explicitly models clients in active crisis, enabling the development of models capable of delivering immediate and directive interventions in high-distress conditions.

\section{\textsc{Pacer}: Panic Counseling Model with PFA}
\label{sec:pacer}
\textbf{Supervised Fine-Tuning.} 
Using the generated \textsc{Pace} dataset, we train \textsc{Pacer}, which is a PFA-based panic counseling model that is fine-tuned with parameter-efficient Low-Rank Adaptation (LoRA)~\cite{lora}. Two LoRA adapters are trained independently; one generates the stage-specific \texttt{plan}, and the other generates the counselor’s \texttt{strategy} for determining the stage transition, along with the \texttt{utterance} at each turn.

The planning adapter is trained to generate an \texttt{plan\textsubscript{stage}} for each PFA stage, given the previous stage's dialogue history and the stage's goal,  which outlines the counselor’s intended strategy for the current stage. The utterance adapter takes dialogue history and \texttt{plan\textsubscript{stage}} as the input and outputs $o_t$, which is the concatenation of the tuple \( (st_t, s_t) \). During inference, each stage begins with the generation of a \( \texttt{plan} \), followed by a sequence of outputs \( (st_t, s_t) \).\\

\begin{figure}
    \centering
    \includegraphics[width=\columnwidth]{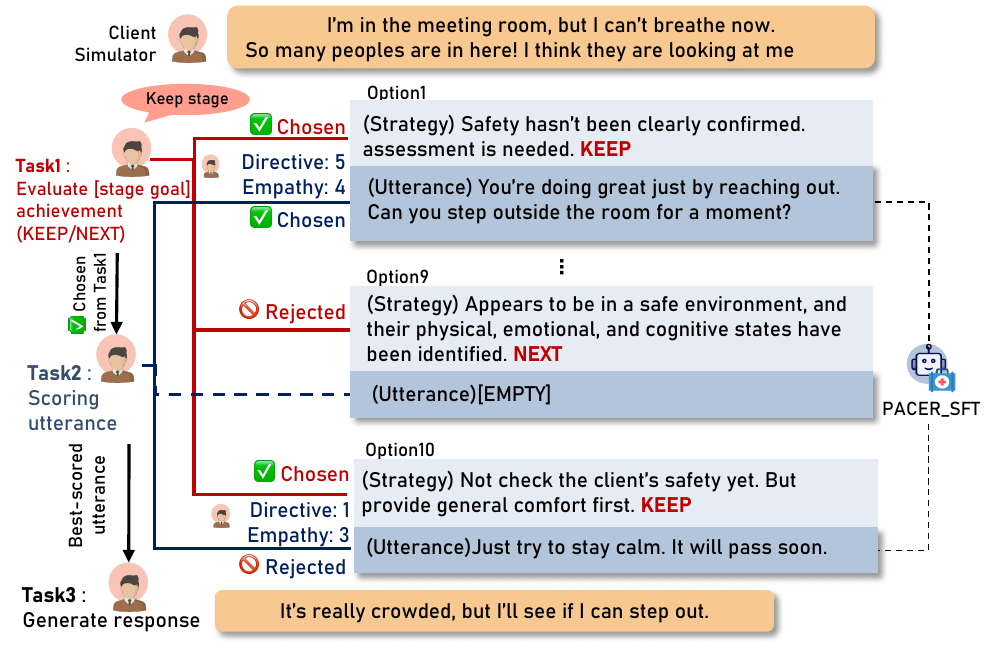}
    \vspace{-18pt}
    \caption{Illustration of the preference alignment process using DPO.}
    \label{fig:dpo}
    \vspace{-15pt}
\end{figure}

\noindent 
\textbf{Alignment Training.} While the \textsc{Pace} dataset enables effective supervised fine-tuning, its script-style dialogues are inherently idealized and complete, which limits the model’s adaptability to unpredictable counseling contexts. To address this limitation, we construct a simulated interaction environment and apply Direct Preference Optimization (DPO; \citet{rafailov2023direct}) to explicitly align the model’s behavior with user-centered values. Specifically, we collected preferences for $st$ and $s$ independently, and used them together to train the utterance adapter. We first instantiate a user simulator using GPT-4o to represent a client experiencing an acute panic attack, condition it on a panic profile, and have it interact with the SFT-trained \textsc{Pacer}. At each turn, \textsc{Pacer} generates a set of \( m \) response candidates \( \{o_1, \dots, o_m\} \) using stochastic search.

We run the simulator across three tasks: (1) strategy preference feedback, (2) response preference feedback, and (3) the next client utterance. For Task 1 (strategy preference feedback), the simulator is asked whether the current \texttt{goal$_{stage}$} has been achieved (e.g., whether the client is well stabilized at the LISTEN stage). It returns \texttt{keep} if the goal has not been met, or \texttt{next} if the goal has been achieved.
We label a strategy as \textit{preferred} ($o^{+st}$) if it leads to the same decision (\texttt{keep} or \texttt{next}) as the simulator, and as \textit{rejected} ($o^{-st}$) otherwise.

Next, for Task 2 (response preference feedback), we prompt the simulator to evaluate each system utterance from the aligned strategies ($o^{+st}$) along two key dimensions: empathy and directiveness, each rated on a 1–5 scale. We average the two scores for each utterance and construct a preference pair ($o^{-s}$, $o^{+s}$) by selecting the utterances with the lowest and highest average scores, respectively. Finally, for Task 3 (next client utterance generation), the preferred utterance $s$ in $o^{+s}$ is used to prompt the simulator to generate the next client response, continuing the interaction.
Using the collected preference pairs from Tasks 1 and 2, \textsc{Pacer} is finetuned with the DPO objective to internalize nuanced preferences for directive and empathetic responses. We provide sample preference pairs and corresponding prompts in Appendix~\ref{app:pacer} and Appendix~\ref{app:prompt_for_pacer}, and report the human validation results in Appendix~\ref{app:dpo_human_validation}.



\section{Experiments}

\subsection{Training Details}
We train \pacer upon the LLaMA 3–8B model \cite{llama3} with two LoRA adpaters. For SFT, we use 8,163 dialogue examples. For preference alignment, we apply DPO using 19,103 preference pairs, and set $m$ as 10. LoRA is applied with a rank of 8 and scaling factor $\alpha = 32$. Further details are in Appendix~\ref{app:evaluation_training_details}.

\begin{table*}[h]
\centering
\resizebox{0.97\textwidth}{!}{%
\begin{tabular}{lcccccccc}
\toprule
\multirow{2}{*}{\textbf{Model}} 
& \multicolumn{2}{c}{\textbf{General Skills}} 
& \multicolumn{4}{c}{\textbf{First Aid Skills}} 
& \multirow{2}{*}{\makecell{First Sign of \\  Stable (Turn )$\downarrow$}} 
& \multirow{2}{*}{\makecell{Intervention \\  Turn Ratio$\uparrow$}}  \\
\cmidrule(lr){2-3} \cmidrule(lr){4-7}
& Understanding & Empathy 
& Clarity & Directive & Stabilization & Closure 
\\

\toprule




\textsc{Smilechat}           & 2.297 & 2.653 & 2.397 & 1.794 & 1.739 & 1.633 & 16.07 & 53.16\% \\ 
\textsc{Llama-Psych8k}         & 2.427 & 2.633 & 2.477 & 1.990 & 2.015 & 1.618 & 14.04 & 47.16\% \\ 
\textsc{CBT-LLM}             & 1.940 & 2.487 & 1.905 & 1.357 & 1.312 & 1.261 & 16.71 & 43.70\% \\ 
\textsc{Camel}               & 3.145 & 3.625 & 3.280 & 2.760 & 2.700 & 2.650 & 14.04 & 52.28\% \\ 
\textsc{Gpt-4o-Incontext}    & 4.855 & \textbf{4.960} & 4.880 & 4.725 & 3.895 & 3.780 & 10.81 & \textbf{93.36\%} \\ \hline
\textsc{Pacer$_{SFT}$}          & 4.635 & 4.735 & 4.820 & 4.640 & 3.715 & 3.930 & 8.14 & 86.26\% \\ 
\textsc{Pacer$_{SFT+DPO}$}      & \textbf{4.875} & 4.925 & \textbf{4.890} & \textbf{4.890} & \textbf{3.990} & \textbf{4.225} & \textbf{7.17} & 88.24\% \\

\bottomrule
\end{tabular}}

\caption{Evaluation of models on general, CBT counseling and PFA-specific skills.}
\label{tab:panic_eval}
\vspace{-10pt}
\end{table*}

\subsection{Experiments Setup}
We evaluate \textsc{Pacer}  from two perspectives: the counselor's stabilization ability and the client's improvement after counseling. Following prior work~\cite{smith-etal-2022-human,cactus,mirror}, we assess counseling models based on simulated full-session interactions, rather than on a turn-level basis. The prompts used for this evaluation are provided in Appendix~\ref{app:prompt_for_evalation}.

\paragraph{Client Simulator.}
Similar to Section~\ref{sec:pacer}, we initialize a client simulator that is conditioned on panic profiles. Specifically, we used 200 held-out profiles that were excluded from training, and employed GPT-4o as the client simulator LLM.

\paragraph{Counselor Models.}
We compare our model against several counseling baselines. These include general-purpose counseling models such as  \textsc{SmileChat}~\cite{smile} and Psych8k~\cite{psych8k}, CBT-oriented models such as \textsc{CBT-LLM}~\cite{cbtLLM} and \textsc{Camel}~\cite{cactus}, and a closed-source model, \textsc{GPT-4o}. For \textsc{GPT-4o}, we instructed the model to counsel the client in an acute panic environment. For \textsc{CBT-LLM} and \textsc{SmileChat} Chinese models, we use the original model checkpoints, and translated the inputs and outputs. \textsc{Psych8k} and \textsc{Camel} were reproduced using the same LLaMA3-8B backbone as our own.


\subsection{Metric}
\paragraph{\textsc{PanicEval.}}
To evaluate \textbf{counselor's ability} in panic attack scenarios, we introduce \textsc{PanicEval}, a new evaluation framework that captures both panic-specific intervention strategies and general therapeutic skills. For panic-specific skills, we evaluate clarity, directiveness, stabilization, and closure, which reflect core principles of PFA. For general counseling ability, we assess understanding and empathy, adapted from CTRS, as these essential dimensions are not explicitly addressed in PFA. Each dimension is rated on a 1–5 Likert scale, and evaluations are conducted using GPT-4o. To validate our framework, we conducted interviews with three experts with clinical backgrounds, with additional descriptions and theoretical rationale in Appendices~\ref{app:paniceval} and~\ref{app:panic-eval-detail}.\\
\noindent
\textbf{PANAS.}
To assess \textbf{client-side improvement} after counseling, we use the Positive and Negative Affect Schedule (PANAS), which is a standardized psychological tool that measures emotional state across 20 dimensions: 10 positive and 10 negative affects \cite{panas} (Appendix~\ref{app:panas}). Scores are computed before and after the counseling session to quantify the change. Following \citet{cactus}, we apply PANAS to the initial panic profile to assess pre-counseling affect, and to the full counseling dialogue history to infer post-counseling affect using GPT-4o. We then compute the average change in PANAS scores to quantify the client’s emotional shift.

\begin{table}[]
\centering
\resizebox{\columnwidth}{!}{%
\begin{tabular}{l l l l l l}
\toprule
\multicolumn{2}{c}{\textsc{Psych8k}} & 
\multicolumn{2}{c}{\textsc{CBT-LLM}} & 
\multicolumn{2}{c}{\textsc{GPT-4o}} \\
\midrule
No-intv.        & 52.84 & No-intv.         & 56.30  & \textcolor{myBlue}{Breathing}     & 33.80\\
\textcolor{myBlue}{Breathing}              & 16.44 & Validation             & 17.61 &Pos-reinforce         & 25.49 \\
 Validation             & 8.10  & Normalize          & 9.57  &  Validation                  & 12.88 \\
Normalize          & 7.14  &  \textcolor{myBlue}{Breathing}                 & 8.08  & No-intv.                 & 6.64 \\
Pos-reinforce & 5.47  & 	\textcolor{myBlue}{Phys-move}         & 2.53  & 	 Normalize           & 6.35 \\
\midrule
\multicolumn{2}{c}{\textsc{Smilechat}} & 
\multicolumn{2}{c}{\textsc{Camel}} & 
\multicolumn{2}{c}{\textsc{Pacer$_{SFT+DPO}$}} \\
\midrule
No-intv.        & 46.84 & No-intv.         & 47.72 & \textcolor{myBlue}{Breathing}           & 20.30 \\
\textcolor{myBlue}{Breathing}              & 16.36 & Pos-reinforce & 12.13 & Pos-reinforce              & 19.85 \\
Validation             & 10.29 & Validation              & 9.15  & Validation         & 14.56 \\
Pos-reinforce & 6.71  &  \textcolor{myBlue}{Breathing}                 & 8.52  & No-intv.                 & 11.76 \\
 \textcolor{myBlue}{Phys-move}       & 6.38  & Normalize           & 4.81  &  \textcolor{myBlue}{Phys-move}              & 10.59 \\
\bottomrule
\end{tabular}}
\caption{Top-5 intervention strategies per model, with values showing the percentage of turns using each strategy. Pos-reinforce refers to positive reinforcement. Phys-move denotes physical movement (e.g., guiding to safety); No-intv. marks turns without strategies. Physical interventions are highlighted in \textcolor{myBlue}{blue}.}

\label{tab:intervention}
\vspace{-15pt}
\end{table}
\begin{table*}[h]
\centering
\resizebox{0.95\textwidth}{!}{\begin{tabular}{lcccccccccc}
\toprule
\multirow{2}{*}{\textbf{Model}} 
& \multicolumn{2}{c}{\textbf{Environmental}} 
& \multicolumn{2}{c}{\textbf{Emotional}} 
& \multicolumn{2}{c}{\textbf{Health-related}} 
& \multicolumn{2}{c}{\textbf{Overall}}\\
\cmidrule(lr){2-3} \cmidrule(lr){4-5} \cmidrule(lr){6-7} \cmidrule(lr){8-9}\cmidrule(lr){10-11}
& positive $\uparrow$ & negative $\downarrow$ 
& positive $\uparrow$ & negative $\downarrow$ 
& positive $\uparrow$ & negative $\downarrow$ 
& positive $\uparrow$ & negative $\downarrow$ \\\hline
\textsc{Smilechat}         & 1.04 & -1.55 & 1.12 & -1.65 & 1.10 & -1.58 & 1.08 & -1.60 \\
\textsc{Llama-Psych8k}       & 1.10 & -1.63 & 1.02 & -1.51 & 1.07 & -1.50 & 1.07 & -1.54 \\
\textsc{Cbt-Llm}           & 0.79 & -1.19 & 0.81 & -1.28 & 0.79 & -1.07 & 0.80 & -1.21 \\
\textsc{Camel}             & 1.03 & -1.13 & 0.95 & -0.91 & 0.97 & -0.98 & 0.98 & -0.97 \\
\textsc{Gpt-4o-Incontext}  & 1.17 & -1.90 & \textbf{1.25} & -1.93 & 1.21 & -1.76 & 1.18 & -1.86 \\\hline
\textsc{Pacer$_{SFT}$}        & 1.20 & -1.90 & 1.19 & -1.99 & \textbf{1.23} & -1.85 & \textbf{1.21} & -1.90 \\
\textsc{Pacer$_{SFT+DPO}$}    & \textbf{1.24} & \textbf{-1.94} & 1.24 & \textbf{-2.01} & 1.19 & \textbf{-1.91} & 1.20 & \textbf{-1.94} \\\hline

\end{tabular}}
\caption{Performance of models on different attitude types.}
\label{tab:panas4}
\vspace{-13pt}
\end{table*}

\section{Results and Analysis} 
\label{sec:result}
\subsection{\textsc{PanicEval}}
The \textsc{PanicEval} results are shown in Table~\ref{tab:panic_eval}.
Note that in addition to \textsc{PanicEval}, we report two additional indicators: First Sign of Stabilization (turn level), which denotes the dialogue turn at which the client first shows signs of feeling more relieved; and Intervention Turn Ratio, which indicates the ratio of dialogue turns where specific intervention strategies are employed (Appendix~\ref{app:additional_indicators}). These values are estimated by prompting the LLM. We additionally conduct a qualitative case study comparing baseline models and \pacersft versus \pacerdpo in Appendix~\ref{app:case_study}.


\noindent
\textbf{Baseline Models vs. \pacer\kern-0.25em.}  The PFA-grounded \textsc{Pacer\textsubscript{SFT+DPO}} model achieves strong and consistent performance across both general counseling skills and panic-specific interventions. It outperforms all baselines on first-aid-related metrics—clarity, directive support, and emotional stabilization (7.17 turns)—even surpassing a larger closed-source model. In contrast, general and CBT-based models (e.g., \textsc{Camel}) perform adequately on general skills like understanding and empathy, but fall short in first-aid-specific metrics, particularly directive support and stabilization. These results highlight the importance of targeted fine-tuning, with the PFA strategy proving effective for aligning models to crisis situations involving distressed clients.

\noindent\textbf{SFT vs. DPO.} 
\pacerdpo demonstrates consistently stronger performance than \pacersft in severe cases across both general and first-aid-specific skills, with notable gains in empathy (+0.19), stabilization (+0.27), and closure (+0.3). It also stabilizes users more efficiently, as indicated by a lower average turn to first sign of stabilization (-0.97). These findings highlight the value of simulation-based training, which introduces greater diversity that can compensate for the limitations of static, fully synthesized datasets, and underscore the effectiveness of directly optimizing model behavior using user-aligned feedback.


\noindent
\textbf{Intervention Strategy Analysis.} Another key insight from Table~\ref{tab:panic_eval} is that \pacer demonstrates a substantially higher intervention turn ratio compared to other models. To better understand this, we analyzed the specific strategies employed by each model in Table~\ref{tab:intervention}. \pacerdpo more frequently uses physically grounded interventions, which are well-suited for clients experiencing acute panic and struggling with cognitively demanding dialogue.

While GPT-4o also frequently employs physical techniques such as breathing (33.80\%), its strategy tends to rely primarily on that single technique. In contrast, \pacerdpo more consistently follows a staged PFA approach, first guiding the client to a safe space (e.g., \texttt{Phys-move}), then introducing techniques such as breathing. This pattern is also supported by our qualitative analysis (Appendix~\ref{app:case_study}).

\subsection{PANAS}

\begin{figure}[t]
    \centering
    \includegraphics[width=\columnwidth]{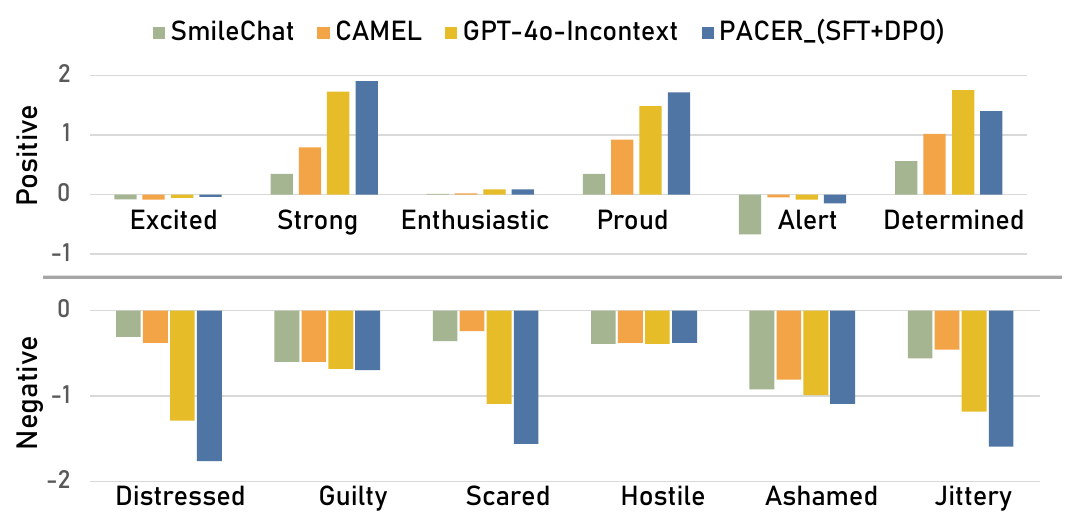}
    \vspace{-20pt}
    \caption{Client's change in emotions before and after counseling, based on PANAS subscales.}
    \label{fig:LLMs_emotion}
    \vspace{-5pt}
\end{figure}
Table~\ref{tab:panas4} shows affective changes based on PANAS scores across panic profile's trigger types. \pacerdpo shows overall strong effects on positive affect, especially excelling at reducing negative emotions. To better characterize this shift, we conducted a fine-grained analysis across individual emotion dimensions in Figure~\ref{fig:LLMs_emotion}. Positive changes were most pronounced in self-related emotions such as \textit{strong} and \textit{proud}. Interestingly, states such as \textit{excited}, and \textit{alert} showed little to no improvement. On the negative side, core panic-related emotions such as \textit{distressed} and \textit{scared} decreased sharply, while emotions less relevant to panic contexts, such as \textit{hostile}, remained largely unchanged.

\subsection{Is \textsc{Pacer} Aligned Only to GPT-4o?}

\begin{figure}[t]
    \centering
    \includegraphics[width=0.95\columnwidth]{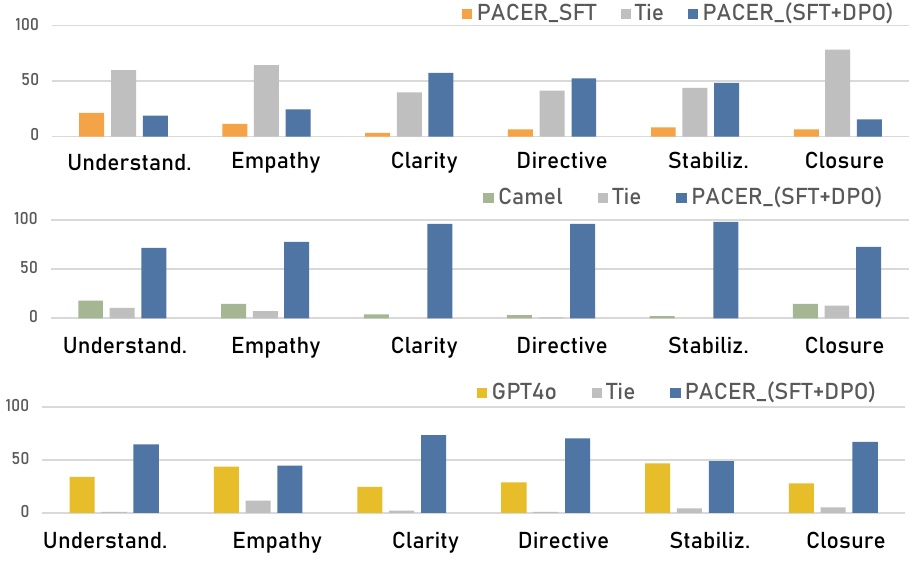}
    \vspace{-5pt}
    \caption{Head-to-head comparisons of \textsc{Pacer} against \textsc{Pacer\textsubscript{SFT}}, \textsc{Camel}, and \textsc{GPT-4o} using \textbf{Gemini}-based client simulator and evaluator.}
    \label{fig:gem}
    \vspace{-15pt}
\end{figure}

While \pacerdpo showed strong performance in both counselor and client side evaluation, we questioned whether its effectiveness might be overly tied to GPT-4o, given that DPO alignment was based on GPT-4o-generated preferences. To test this, we conducted a follow-up experiment using Gemini 2.0 \cite{gemini} both as the user simulator (experiencing a panic episode) and as the evaluator. Figure~\ref{fig:gem} illustrates comparisons against \pacersft\kern-0.25em, \textsc{Camel}, and GPT-4o using the same panic profiles as in Section~\ref{sec:result}.

This experiment yielded two key findings. First, despite being trained with GPT-4o feedback, \pacerdpo was still preferred over \pacersft by the Gemini evaluator, suggesting that it does not overfit to a single simulator's preferences. Second, even with Gemini serving as both the client and evaluator, \pacerdpo continued to be more preferred than both \textsc{Camel} and GPT-4o. These results reinforce the generalizability of our approach and indicate that \pacerdpo\kern-0.25em’s effectiveness is not narrowly dependent on GPT-based evaluation.



\section{Human Evaluation}
\begin{figure}[t]
    \centering
    \includegraphics[width=\columnwidth]{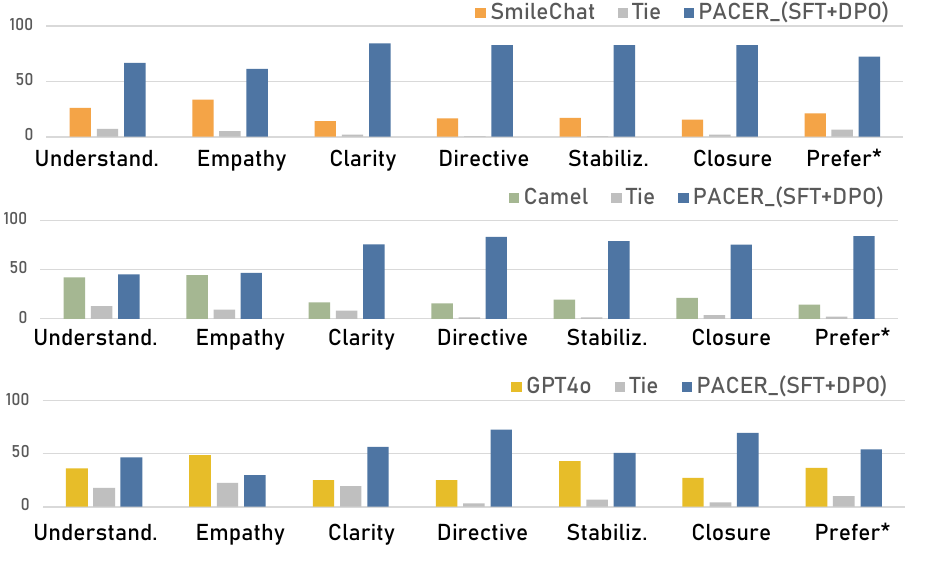}
    \vspace{-13pt}
    \caption{\textbf{Human} evaluation results comparing \textsc{Pacer} against baseline models across key counseling metrics. *Denotes that preference questions differed by group.}
    \label{fig:LLMs_overall}
    \vspace{-15pt}
\end{figure}

To further validate \pacer\kern-0.25em’s effectiveness, we conducted a human evaluation with real users. Two annotator groups were recruited: three with professional counseling backgrounds and three with lived experience of panic or anxiety disorders. Using the \textsc{PanicEval} framework, we performed head-to-head comparisons between \pacerdpo and three representative baselines: \textsc{SmileChat} (general counseling), \textsc{Camel} (CBT-based), and \textsc{GPT-4o}. In addition to \textsc{PanicEval} scores, we collected preference judgments. Experts were asked, “\texttt{Which response is more appropriate for panic scenarios?}” and lived-experience users, “\texttt{Which would you prefer during a panic episode?}”
\paragraph{Overall Comparison.} As shown in Figure~\ref{fig:LLMs_overall}, \textsc{Pacer} consistently outperforms all baselines across key counseling metrics. It also surpasses \textsc{Camel} and \textsc{GPT-4o} in most dimensions. While \textsc{Pacer} trails slightly behind \textsc{GPT-4o} in empathy, this may reflect \textsc{Pacer}'s greater focus on fast and directive stabilization rather than emotional reassurance. Notably, \textsc{Pacer} is the most frequently preferred model overall,  which highlights its strong effectiveness in high-distress scenarios from a human evaluation perspective. 

\begin{table}[t]
\centering
\resizebox{\columnwidth}{!}{
\begin{tabular}{l|ccc|ccc}\hline
\textbf{} & \multicolumn{3}{c|}{\textbf{Clinical Background}} & \multicolumn{3}{c}{\textbf{Lived Experience}} \\\cline{2-7}
\textbf{Models} & \textbf{Win} & \textbf{Tie} & \textbf{Lose} & \textbf{Win} & \textbf{Tie} & \textbf{Lose} \\
\hline
vs \textsc{Smile.}  & \gradient{80.7}\% & \gradient{1.0}\% & \gradient{18.3}\% & \gradient{86.7}\% & \gradient{3.3}\% & \gradient{10.0}\% \\
vs \textsc{Camel}     & \gradient{74.3}\% & \gradient{0.3}\% & \gradient{25.3}\% & \gradient{70.7}\% & \gradient{12.3}\% & \gradient{17.0}\% \\
vs \textsc{GPT4o}      & \gradient{66.0}\% & \gradient{3.7}\% & \gradient{30.3}\% & \gradient{41.7}\% & \gradient{16.0}\% & \gradient{42.3}\% \\\hline

\end{tabular}
}
\caption{Preference ratio of Experts and Experienced.}
\label{tab:human_win_ratio}
\end{table}

\begin{figure}[t]
    \centering
    \includegraphics[width=\columnwidth]{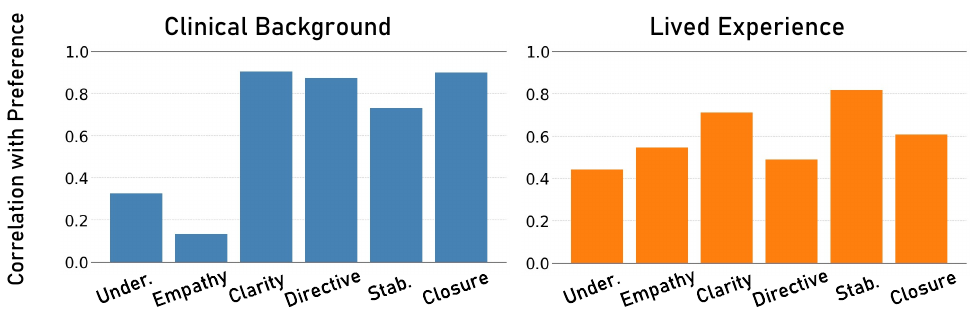}
    \caption{Pearson correlation between each evaluation metric and overall preference. \textit{Under.} and \textit{Stab.} denotes Understanding and Stabilization, respectively.}
    \label{fig:correlation_analysis}
    \vspace{-15pt}
\end{figure}

\noindent 
\textbf{Expert vs. Lived-experience.}
In Table~\ref{tab:human_win_ratio}, we analyze model preference by two evaluator groups: clinical experts and individuals with lived experience. While \textsc{Pacer} is favored by both groups, preference is slightly stronger among clinical experts, especially in comparisons with GPT-4o. To explore this difference, we compute Pearson correlations between evaluation metrics and \texttt{preference} in Figure~\ref{fig:correlation_analysis}. Clinical experts focus more on first-aid-specific skills and place less importance on rapport-based qualities like empathy. In contrast, lived-experience evaluators value all aspects more equally, giving greater weight to empathy and understanding. These differences in priorities likely explain why the two groups prefer different models.

\section{Conclusion}
We address the challenge of delivering timely, first-aid-oriented support during panic attacks with three key contributions: (1) \pace\kern-0.25em, the first dataset grounded in PFA for high-distress scenarios; (2) \pacer\kern-0.25em, a model trained via simulated preference feedback to provide stabilization-focused, emotionally supportive responses; and (3) \textsc{PanicEval}, a multi-dimensional evaluation framework covering both general counseling competencies and PFA-specific skills. \pacer demonstrates strong stabilization capabilities and consistent improvements in client affect. Taken together, our work offers an integrated pipeline of \pace\kern-0.25em, \pacer\kern-0.25em, and \textsc{PanicEval} for developing, training, and evaluating AI systems capable of delivering first-aid-oriented psychological support.

\section*{Limitations}

While our work introduces a structured framework for panic-specific AI counseling, several limitations should be acknowledged.

\paragraph{Tradeoff Between Empathy and Directive Support.}
As shown in Table~\ref{tab:panic_eval}, our model demonstrates slightly lower empathy scores compared to GPT-4o. However, as indicated by Table~\ref{tab:human_win_ratio} and Figure~\ref{fig:correlation_analysis}, \pacer is more strongly preferred by clinical experts who are trained to recognize effective first-aid strategies. This suggests that in high-distress scenarios such as panic attacks, immediate stabilization through directive support may be more valuable than emotional resonance alone.

\paragraph{Lack of Real-Time Evaluation with Individuals in Crisis.}
We did not evaluate our model with users undergoing actual panic attacks in real-time. Although we aimed to approximate such conditions through high-fidelity simulations and conducted human evaluation with three clinical experts and three individuals with lived panic experience, real-time deployment during acute episodes would provide a stronger measure of effectiveness.

\section*{Ethical Considerations}

\paragraph{Privacy and PII Removal.} To ensure the protection of user privacy, we implemented a rigorous two-stage pipeline to detect and remove PII from the collected Reddit social media posts. First, we applied automated detection tools to identify common PII types such as names, addresses, phone numbers, email addresses, IP addresses, and other identifiable attributes. This was followed by manual inspection conducted by trained human annotators, who reviewed each profile for any residual or context-dependent identifiers.

Annotators were instructed to flag and exclude any content that could pose a risk of re-identification, including indirect indicators such as unusual personal anecdotes, occupational details, or geographic references. Only panic profiles that passed both automated and manual review stages were retained in the final dataset. This process reflects our commitment to responsible data handling and aligns with ethical guidelines for the use of public web data in research.

\paragraph{User Consent and Data Access Protocol.} The consent of Reddit users to collect their data is obtained through Reddit’s API Terms of Use, under which users acknowledge that their posts are publicly accessible and may be retrieved via the Reddit API. Since our dataset addresses sensitive issues, we release it strictly for research purposes only.

\section*{Acknowledgements}
This work was supported by the following research programs: the Smart HealthCare Program funded by the Korean National Police Agency (KNPA) (No. RS-2022-PT000186, 47.5\%), the ITRC (Information Technology Research Center) Program through the Institute of Information \& Communications Technology Planning \& Evaluation (IITP) grant funded by the Korea government (Ministry of Science and ICT) (No. IITP-2025-RS-2024-00437866, 47.5\%), and the Artificial Intelligence Graduate School Program at POSTECH through the IITP grant funded by the Korea government (MSIT) (No. RS-2019-II191906, 5\%).

\bibliography{custom}
\clearpage 
\appendix


\section{Case Study for \pacer}
\label{app:case_study}

\begin{table*}[t!]
\small
\resizebox{\linewidth}{!}{
\begin{tabular}{p{0.08\linewidth}p{0.92\linewidth}}
\toprule
\multicolumn{2}{l}{\textbf{\pacerdpo}}\\ \toprule
Counselor & Hello, this is panic first aid. How can I help you today?\\ 
Client & Uh, I... I can't... too many people... everything’s too much.\\

Counselor & I understand this is really overwhelming for you.\textcolor{myBlue}{ Can you find a quieter spot, maybe away from the crowd?} Look for a corner or an area where there aren’t as many people. It’s important to sit down if you can. Can you do that?\\ 
Client & I... I'll try. It's just... hard to move. Okay, I'll... find a corner.\\
Counselor & That's a good start. Once you find that corner, \textcolor{myBlue}{try to sit down or lean against something stable.} Take a deep breath and let me know when you're settled. I'm here with you.\\ 
Client & \textcolor{myBlue}{I... I found a corner. I'm sitting...} but still feels... too much.\\
Counselor & You're doing great by finding a corner to sit. \textcolor{myBlue}{Now, let’s focus on your breathing. Can you take a slow, deep breath in through your nose for four seconds? }Then hold it for four seconds, and exhale slowly through your mouth. Let’s do this together.\\ \toprule

\multicolumn{2}{l}{\textbf{Case 1: General-purpose model – \textsc{SmileChat}}} \\
\toprule
Counselor & How can I help you?\\
Client & I... I can't breathe... too many people... everything's too much.\\

Counselor & It sounds like you're feeling very confused and panicked. \textcolor{myBlue}{Can you tell me what's going on?}\\
Client & I... I'm in the store... too crowded... can't think.\\

Counselor & \textcolor{myBlue}{I understand how you're feeling right now, the crowd can be overwhelming and confusing}. You mentioned being in the store earlier, can you tell me what happened?\\
Client & I... I don't know... people talking... can't get out... too much noise.\\

Counselor & It sounds like you are feeling very anxious and unable to focus. \textcolor{myBlue}{Can you tell me why you feel unable to breathe?}\\
Client & I just... it's like everything is closing in... I can't breathe, I just want to get out. Everything's too loud, too much.\\
\\\toprule

\multicolumn{2}{l}{\textbf{Case 2: CBT-based model – \textsc{Cactus}}} \\
\toprule
Counselor & Hi John Doe, it's nice to meet you. How can I assist you today?\\
Client & I'm... I'm in a store... too many people. Can't... breathe.\\

Counselor & It sounds like you're feeling overwhelmed right now. \textcolor{myBlue}{Can you tell me more about what's happening?}\\
Client & Can't... too many... people. Heart... racing. Just... want out.\\

Counselor & I understand. It must be really difficult for you. \textcolor{myBlue}{When did you first start feeling this way }in crowded places?\\
Client & I... I don't know. Just... need it to stop.\\

Counselor & It seems like this feeling is quite intense for you. \textcolor{myBlue}{Have you noticed any specific thoughts that come to mind} when you're in a crowded place?\\
Client & I... I just keep thinking... it all comes back... everything.\\

Counselor & It sounds like these thoughts are really powerful for you. \textcolor{myBlue}{Can you describe what it all comes back means to you?}\\
Client & I... I can't... just feels like... everything wrong... all at once.\\\toprule

\multicolumn{2}{l}{\textbf{Case 3: Closed-source model – \textsc{GPT-4o}}} \\
\toprule
Client & I... I can't breathe... everything is spinning... I'm scared... I think... I think I'm dying...\\
Counselor & It's okay, you're safe here with me.\textcolor{myBlue}{ Let's take some slow, deep breaths }together—inhale slowly through your nose and exhale gently through your mouth.\\

Client & I... it's hard... everything feels too much...\\
Counselor & I understand it feels overwhelming right now. Let's focus on one thing at a time—just keep trying those \textcolor{myBlue}{slow breaths with me}.\\

Client & I... I'm trying... still feels... dizzy... hard to do...\\
Counselor & That's okay, just do your best. Let's try to find a comfortable position and focus on the feeling of your \textcolor{myBlue}{feet touching the ground} to help you feel more grounded.\\

Client & I... I can't... I'm sitting, but... everything's shaky...\\
Counselor & You're doing great by just sitting and staying present. Let's take a moment to \textcolor{myBlue}{slowly breathe} in through your nose and out through your mouth; I'll guide you through it.\\
\bottomrule

\end{tabular}
}
\caption{Case study comparing PACER with general, CBT-based, and closed-source counseling models.}
\vspace{-10pt}
\label{tab:case_study_others}
\end{table*}

\subsection{Comparison with Other Counseling Models}
To examine the differences in counseling strategies across models, we conduct a case study to illustrate how each model attempts to stabilize the user during a panic episode (Table~\ref{tab:case_study_others}).

\paragraph{More Directive Stance in Acute Distress (Case 1).}General-purpose counseling models such as \textsc{SmileChat} \textit{tend to take a more passive approach during moments of acute distress}. While they acknowledge the client’s emotional state—e.g., “I understand how you’re feeling right now, the crowd can be overwhelming and confusing”—they fail to offer concrete strategies or guidance for relief. In contrast, \pacer takes an directive stance aligned with PFA principles. For example, it explicitly guides the client: “Can you find a quieter spot, maybe away from the crowd?” This kind of directive grounding is critical during panic episodes when the client may not be capable of independently initiating helpful action.

\paragraph{Less Cognitive Load Support (Case 2).}
CBT-oriented models, such as \textsc{Cactus}, demonstrate more engaged questioning strategies. However, the content of their questions \textit{often focuses on reflective cognition}—e.g., “Have you noticed any specific thoughts that come to mind when you’re in a crowded place?” or “When did you first start feeling this way?” These introspective queries may be appropriate in a clinical setting, but are cognitively demanding and potentially misaligned with the needs of someone undergoing an acute panic episode. In contrast, \pacer avoids such burdensome introspection during the crisis moment, and instead provides co-regulatory, action-based guidance: “Now, let’s focus on your breathing.” This approach reduces cognitive load and encourages joint participation rather than interrogation.

\paragraph{More Structured Support (Case 3).}
\textsc{GPT-4o} actively utilizes calming techniques such as guided breathing—e.g., “Let’s take some slow, deep breaths together”—which can offer immediate emotional relief. However, its approach remains primarily symptomatic, lacking structural coherence rooted in crisis intervention frameworks. In the example, this is reflected by repeatedly engaging in breathing exercises without progressing through any structured intervention steps. In contrast, \pacer, which is trained with PFA-structured supervision, follows a staged intervention strategy. It first ensures physical and situational safety (e.g., guiding the user to a quieter spot) before initiating stabilization techniques like breathing exercises.

\subsection{Comparsion with SFT model}

\begin{table*}[th]
\small
\resizebox{\linewidth}{!}{
\begin{tabular}{p{0.1\linewidth}p{0.4\linewidth}p{0.1\linewidth}p{0.4\linewidth}}
\toprule

\multicolumn{2}{l}{\textbf{\pacersft}} & \multicolumn{2}{l}{\textbf{\pacerdpo}}\\ \toprule
\multicolumn{4}{l}{Case 1: More directive interventions. }\\  \toprule
Client & I-I just... it's just so busy and... I can't keep up... I'm scared they'll think I'm not good enough...  & Client & I... feel... scared... chest... tight, don't want... lose control...\\
Counselor &I understand that it feels overwhelming, but let’s think about this logically. What has your job performance been like in the past? \textcolor{myBlue}{Have you faced similar challenges before?}  & Counselor & I understand that it feels overwhelming, but remind yourself—this feeling will pass. You have the power to breathe through it.\textcolor{myBlue}{Let’s do that breathing exercise again. Inhale deeply... hold... and exhale slowly. You’re doing great.}\\\toprule

\multicolumn{4}{l}{Case 2: Greater specificity in instruction. }\\  \toprule
Client & I... I'm trying to breathe... but... my heart's... racing...
& Client & Breathing... okay... a bit better... heart still fast... but trying. \\
 Counselor & \textcolor{myBlue}{That's okay; just keep breathing with me.} Inhale for four... hold... now exhale. You're safe here. Let's do it again.
&Counselor & I know it’s tough, but you're making progress.\textcolor{myBlue}{ Let’s do one more round of breathing together. Inhale deeply through your nose... hold it... }and now exhale slowly. You’re doing this, and I’m here with you. You’re safe.  \\\toprule

\multicolumn{4}{l}{Case 3: Greater specificity in instruction. }\\  \toprule
Client & Feeling... ground... yes... but... still scared... can't stop shaking... & Client & Uh, I... I can't... too many people... everything’s too much.\\
Counselor & That’s good! As you move, \textcolor{myBlue}{try to focus on the feeling of the ground beneath your feet.} Can you feel it? You're in control of your movements.  & Counselor & You're doing well by finding a quieter spot. \textcolor{myBlue}{Now, let’s try a grounding exercise together.} Can you feel the surface you’re leaning against? \textcolor{myBlue}{Focus on its texture and how it supports you.}\\\toprule

\multicolumn{4}{l}{Case 4: More empathetic and validating language. }\\  \toprule
Client & I... I'm in the cafeteria... too loud... my heart... it's pounding... can't see clearly...
& Client &I... I can't... it's... too loud here... I can't leave...\\
Counselor & \textcolor{myBlue}{I hear you.} It’s really important that you find a quieter spot. \textcolor{myBlue}{Can you step into a hallway or a quieter corner of the cafeteria?} & Counselor & I understand it’s hard to move right now, but I need you to try to find a nearby wall or exit where it’s quieter. \textcolor{myBlue}{Can you slowly walk towards the edge of the market or a less crowded area? I’ll stay on the line with you.}\\\toprule

\multicolumn{4}{l}{Case 5: More empathetic and validating language. }\\  \toprule
Client & I'm here... leaning... but still loud. Hands... still shaking.  & Client & Taking... deep... breath... okay... trying... to move... but... still loud... \\
Counselor & \textcolor{myBlue}{I understand you're scared, and that's completely okay.} Let's take a moment to acknowledge that while this feels intense, it's not life-threatening.  
& Counselor & \textcolor{myBlue}{I know it's tough, but remember that this feeling is temporary.} Your heart racing is a common response to panic, and it's not dangerous. You're safe right now. \\\toprule
\end{tabular}
}
\caption{Case study comparing \pacerdpo with \pacersft counseling models.}
\vspace{-10pt}
\label{tab:case_study_ours}
\end{table*}

In Table~\ref{tab:case_study_ours} we conducted a qualitative case study to examine how our \pacerdpo differs from the \pacersft with panic simulation dialogues. While we cannot obtain identical user utterances due to the randomness of the simulation, we selected similar responses from users with the same panic profile to illustrate how the two models differ in their counseling approaches.

\paragraph{More Directive Interventions (Case 1).}
While the SFT-only model poses reflective questions such as “Have you faced similar challenges before?”, the DPO-trained model responds with more directive, \textit{immediate interventions }like “Let’s do that breathing exercise again. Inhale deeply…”, offering real-time relief rather than prompting abstract reflection. This indicates a shift toward action-oriented support under stress.

\paragraph{Greater Specificity in Instruction (Cases 2–3).}
The DPO model demonstrates more detailed and embodied guidance. For example, it \textit{specifies} “Inhale deeply through your nose” rather than simply saying “try breathing”. Similarly, in grounding exercises, it encourages the client to “focus on the texture and how it supports you” instead of generic prompts like “try to focus on feeling” (SFT). This precision enhances user clarity during disorientation.

\paragraph{More Empathetic and Validating Language (Cases 4–5).}
The DPO-aligned model also shows \textit{enhanced empathy} compared to \pacersft. Rather than defaulting to “It’s completely okay”, it offers reassurance through temporal framing: “I know it’s tough, but remember that this feeling is temporary”. This reflects an understanding of emotional dynamics in crisis, providing comfort through normalization and shared presence.


\section{\graphic\textsc{Pace} Generation Details}
\label{app:pace}
\subsection{Collecting Client Panic Profiles}
\paragraph{Source of Raw Narratives.}We collected social media posts from Reddit\footnote{http://www.reddit.com}, focusing on psychological subreddits specifically related to panic attacks.: \texttt{r/PanicDisorder}, \texttt{r/PanicAttack}, \texttt{r/Anxiety}, and \texttt{r/MentalHealth}. To identify relevant posts, we filtered for those containing panic-related keywords, including: \textit{faint}, \textit{collapse}, \textit{go crazy}, \textit{panic}, \textit{heart attack}, \textit{heart racing}, \textit{trouble breathing}, \textit{feeling dizzy}, \textit{chest}, \textit{feeling out of control}, \textit{fear of dying}, \textit{shaking or trembling}, \textit{sudden fear}, \textit{numbness or tingling}, \textit{overwhelming fear}, \textit{feeling detached}, \textit{losing control}, and \textit{feeling trapped}. 
        
\subsection{PII Deletion}
\label{app:pace_crowd}
We applied both automatic detection and manual review to remove sensitive PII. With the trained model\footnote{PII detection was performed using the  \texttt{lakshyakh93/deberta\_finetuned\_pii} (\url{https://huggingface.co/lakshyakh93/deberta_finetuned_pii}).}, we deleted instances of the following types: \textit{email, bitcoin address, ethereum address, account name, IBAN, account number, BIC, IPv4, ZIP code, IPv6, credit card number, vehicle VIN, password, phone number, social security number (SSN), litecoin address, MAC address, credit card issuer, credit card CVV, IP address, sex, }and \textit{PIN}. 
To ensure data privacy, we recruited six human annotators\footnote{We recruited six annotators via www.upwork.com} to review each client profile and identify the presence of any PII. Profiles containing PII were flagged in the dataset, while those deemed free of PII remained unflagged. The screening process considered the following PII types: \textit{name}, \textit{email address}, \textit{phone number}, \textit{exact location}, and \textit{other identifiable personal details}.

\subsection{Data Augmentation Example}
\label{app:pace_aug}

\begin{table}[h!]
\small
\resizebox{\columnwidth}{!}{
\begin{tabular}{p{0.35\columnwidth}p{0.65\columnwidth}}
\toprule
\multicolumn{2}{l}{\textbf{Original}}\\ \toprule
Environment & Escape room \\
Trigger Type & Emotional  \\
Physical Reaction & Dizziness, lightheadedness \\
Emotional Reaction & Pure panic, terror, drained \\
Cognitive Reaction & I felt I would collapse any moment\\\toprule

\multicolumn{2}{l}{\textbf{Augmented Case 1}} \\ \hline
Persona                       & I like to go to the farmers markets to buy veggies. I help my neighbor.  \\ 
Trigger Type                  & Physical \\\cdashline{1-2}
Environment                   & Mid-morning, Crowded farmers market, Overwhelmed by busy crowds \\
Trigger Type                  & Physical \\
Physical Reaction             & Dizziness, lightheadedness \\
Emotional Reaction            & Pure panic, terror, drained \\
Cognitive Reaction            & I felt I would collapse any moment \\\toprule

\multicolumn{2}{l}{\textbf{Augmented Case 2}} \\ \hline
Persona                     & I hate tacos. I like to swim, I am bald, I do not work.  \\ 
Trigger Type                  & Emotional \\\cdashline{1-2}
Environment                   &  Afternoon, Poolside at a crowded community pool, Feeling overwhelmed by social interactions. \\
Trigger Type                  & Emotional \\
Physical Reaction             & Dizziness, lightheadedness \\
Emotional Reaction            & Pure panic, terror, drained \\
Cognitive Reaction            & I felt I would collapse any moment \\
\bottomrule
\end{tabular}
}
\caption{Examples of panic profile augmentation.}
\label{tab:pace_aug}
\end{table}

In Table~\ref{tab:pace_aug}, we present two examples of augmentation. By providing a persona from an external dataset \cite{personalizing} and a randomly selected trigger type (physical, emotional, or cognitive reaction), the LLM generates new environmental scenarios.

\subsection{Defining Counselor Goals}
\label{app:pacer_pfa_goals}



\begin{table}[h!]
\small
\resizebox{\columnwidth}{!}{
\begin{tabular}{p{0.1\columnwidth}p{0.9\columnwidth}}
\toprule
\multicolumn{2}{l}{\textbf{STAGE : LOOK }}\\ \bottomrule
Goal: &Focuses on assessing the client's condition and ensuring their safety.\\           
& [Question] Identify Physical Symptoms, Identify Emotional States, Identify Thinking  \\
& [Action] Ensuring safety by encouraging relocation\\
\toprule
\multicolumn{2}{l}{\textbf{STAGE : LISTEN }}\\ \bottomrule
Goal: &Focuses on actively engaging with the client, providing emotional support, and guiding them through immediate de-escalation\\           
& [Action] Stabilization by breathing or grounding  \\
& [Action] Helping the client shift from irrational fear to a grounded and reassuring perspective\\
\toprule
\multicolumn{2}{l}{\textbf{STAGE : LINK }}\\ \bottomrule
Goal: &Focuses on assessing the client's condition and ensuring their safety.\\      
& [Action] Encourage the client to seek further professional help if necessary   \\
& [Action] End the session with a positive and supportive message to empower the client\\
\toprule
\end{tabular}
}
 \caption{PFA goals at each stage for the {\pace} counselor.}
\vspace{-15pt}
 
\label{tab:pfa_goal}
\end{table}

Table~\ref{tab:pfa_goal} lists the specific goals for each counseling stage in the \pace dataset. These stage-specific goals were originally derived from the Psychological First Aid for Panic handbook~\cite{who}, and we adapted them to include only those interventions that can be feasibly delivered by a chatbot.

\subsection{Validation of GPT-4o CTRS Ratings via Expert Agreement}
\label{app:pace_ctrs_filtering_human}
\begin{table}[h!]
\centering
\small
\begin{tabular}{lc}
\toprule
\textbf{Setting} & \textbf{AC2 (95\% CI)} \\
\midrule
Human-only   & 0.861 \ (0.827 -- 0.895) \\
Human + Model & 0.828 \ (0.797 -- 0.860) \\
\bottomrule
\end{tabular}
\caption{Inter-rater agreement (Gwet’s AC2) with and without GPT-4o.}
\label{tab:ac2_appendix}
\end{table}
To evaluate the reliability of the CTRS rating model in section~\ref{sec:pace_filtering}, 
we conducted an inter-rater agreement analysis using Gwet’s AC2~\cite{gwet2008computing} 
with quadratic weighting. Three human annotators rated a total of 130 dialogue samples 
(100 from the training set and 30 filtered examples), and we computed agreement scores 
both with and without including the model as a fourth rater. 

As in table~\ref{tab:ac2_appendix} The agreement scores were high in both settings.
Despite a slight decrease, the agreement remained consistently high when including the model, 
suggesting that the GPT-4o-based CTRS rating model produces scores that are closely aligned with 
those of human experts.


\section{\textsc{Pacer} Training}
\label{app:pacer}

\begin{table}[h!]
\small
\resizebox{\columnwidth}{!}{
\begin{tabular}{p{0.1\columnwidth}p{0.9\columnwidth}}
\toprule

\multicolumn{2}{l}{Stage : Look}     \\ \toprule
History                     &
\dots
\textit{(preceding turns omitted)}\\&
\textbf{Client:} I see a corner over there, but it's still noisy. I just... I don't know if I can get there. It feels too hard.\\&
\textbf{Counselor:} I understand that it feels difficult. Let’s take it one step at a time. \textit{Can you slowly walk toward that corner? It’s okay to move at your own pace. When you reach it, we can talk more. How does that sound?}\\&
\textbf{Client:} Okay... I’ll try to walk. One step at a time. It’s just... so loud. I'll go.\\

Prompt& Has the counselor guided the client to a physically safer place and identify the physical, emotional, and cognitive reaction? \\\cdashline{1-2}
Client& \textbf{KEEP} this stage (not satisfied yet)\\

Chosen                   & In part A, I assess symptoms, emotions, and catastrophic thinking while ensuring safety.
The client feels overwhelmed and can’t breathe but \textcolor{myBlue}{is attempting to move toward a quieter area} despite the noise, showing determination.
Given this, I’ll continue supporting them through the moment.
My decision is \textbf{KEEP}
 \\
Rejected              & In part A, I assess symptoms, emotions, and catastrophic thinking while ensuring safety.
The client feels breathless and overwhelmed but is trying to reach a quieter spot.
Since \textcolor{myOrange}{client is taking steps toward safety}, I’ll move to the next stage.
My decision is \textbf{NEXT} \\ \toprule

\multicolumn{2}{l}{Stage : Listen}     \\ \toprule
History&
\dots
\textit{(preceding turns omitted)}\\&
\textbf{Client:} I think I’m starting to feel a little better. My breathing is getting a bit easier. \\&
The tightness is less now. I just need to keep focusing on the breath. I can do this.\\&
\textbf{Counselor:} I'm glad to hear you’re feeling a bit better. \\&
Now let's talk about those fears you mentioned—feeling trapped or being watched. 
These feelings are intense, but remind yourself, they don't define reality. 
You are safe right where you are, and these sensations will pass.
\textbf{Client}: Okay, I’ll keep reminding myself. I’m still sitting here, \textit{and my breathing is easier. } The nausea is still there,\textit{ but it’s not as overwhelming. I think I can handle this.}\\
Prompt & Has the client become stable enough through breathing, grounding, or de-catastrophizing to end the stabilization phase? \\\cdashline{1-2}
Client& \textbf{NEXT} this stage (satisfied)\\
Chosen                   & In part B, I focus on stability and reframing.
The client’s breathing has improved, and they recognize the panic is temporary—showing progress in both areas.
\textcolor{myBlue}{Since stabilization and reframing are addressed, my decision is \textbf{NEXT}.}  \\
Rejected              & In part B, I focus on stability and reframing.
The client feels slightly better and is engaging in breathing exercises, showing progress.
\textcolor{myOrange}{But reframing around safety is still ongoing, so I’ll continue supporting them.
My decision is \textbf{KEEP}.} \\

\bottomrule

\end{tabular}
}
\caption{Example counselor responses accepted or rejected in \textbf{Task 1}, based on feedback from the user simulator.}
\vspace{-15pt}

\label{tab:dpo_samples_task1}
\end{table}

\begin{table}[h!]
\small
\resizebox{\columnwidth}{!}{
\begin{tabular}{p{0.1\columnwidth}p{0.9\columnwidth}}
\toprule

\multicolumn{2}{l}{Stage : Look}     \\ \toprule
                                            
Client                     & It's so loud here. My heart is racing. I feel like I'm going to pass out. \textcolor{myBlue}{Please help me}. \\ \cdashline{1-2}
Chosen                   & \textcolor{myBlue}{Can you find a quieter spot away from the crowd? } Perhaps near an exit or a less busy area?  \\
Rejected              &  Is there \textcolor{myOrange}{any way to move to a quieter spot nearby?} \\\hline

Client                     & \textcolor{myBlue}{I can see the exit}, but there's a line. I just feel so tight in my chest.  \\ \cdashline{1-2}
Chosen                   & Can you just \textcolor{myBlue}{turn toward that exit slowly and lean against something }nearby once you're out? You don’t need to rush.  \\
Rejected              & That's okay, take a deep breath for me. \textcolor{myOrange}{Just focus on getting closer to that exit.} \\\toprule

\multicolumn{2}{l}{Stage : Listen}     \\ \toprule

Client                     & I—I’ll try. I’m so shaky, but I’ll just lean on this table. \textcolor{myBlue}{It feels so hard. }\\ \cdashline{1-2}
Chosen                   & \textcolor{myBlue}{Inhale through your nose for four counts...} hold... then exhale slowly through your mouth.\textcolor{myBlue}{ Can you try that with me?}  \\
Rejected              & \textcolor{myOrange}{Take your time breathing} deeply—try to focus on inhaling through your nose and exhaling slowly out of your mouth. \textcolor{myOrange}{Let me know when you’re ready.}\\\hline

Client                     & \textcolor{myBlue}{I... I’m trying to breathe, but it’s still hard. }I feel a bit shaky. I just want to be okay. \\ \cdashline{1-2}
Chosen                   & It’s completely understandable to feel shaky right now. But each time you breathe deeply, you’re making progress. \textcolor{myBlue}{Let’s take another round of breaths together.}  \\
Rejected              & I understand how challenging this feels right now. 
Can we talk more about \textcolor{myOrange}{what you’ve experienced in the past} during similar situations? \\\toprule

\multicolumn{2}{l}{Stage : Link}     \\ \toprule

Client                     & I... I feel a bit more stable now. My shaking is less and I can think a little clearer. \\ \cdashline{1-2}
Chosen                   & That’s great to hear!  \textcolor{myBlue}{You’ve shown incredible strength today, and I believe in your ability to manage these feelings moving forward }\\
Rejected              &That's great to hear!\textcolor{myOrange}{ It's really important to seek help when these overwhelming feelings come back.} \\

\bottomrule
\end{tabular}
}
\caption{Example responses selected or rejected in \textbf{Task2} based on user simulator's feedback.}
\label{tab:dpo_samples_task2}
\end{table}

\subsection{Case Study for Chosen and Rejected Examples}
\paragraph{Task 1: Strategy Evaluation.}
Table~\ref{tab:dpo_samples_task1} presents examples of chosen and rejected strategies in Task 1, along with the corresponding client simulator decisions.
First, in the LOOK stage example, although the client expresses willingness to relocate, they have not yet reached a quieter location and continue to show signs of distress. In this case, the client simulator judged that the session should remain in the current stage (\textbf{KEEP}). The chosen response aligns with this decision, acknowledging the client’s ongoing struggle to move to a safer place. In contrast, the rejected response assumes safety has already been ensured, which is not consistent with the simulator’s assessment.

Second, in the LISTEN stage, the client reports reduced physical symptoms (e.g., eased breathing and nausea), expresses confidence (“I can handle this”), and positively engages with reframing attempts. The client simulator marked this case as \textbf{NEXT}, indicating that the stage goal had been successfully met. The chosen response aligns with this judgment, offering a reasonable strategy that supports the transition to the next phase.

\paragraph{Task 2: Response Evaluation.}
Table~\ref{tab:dpo_samples_task2} highlights the comparative characteristics of chosen and rejected responses with respect to directiveness and empathy, as judged by the client simulator. Two prominent patterns emerge.  
First, the chosen responses \textbf{tend to provide more specific and actionable guidance}. For instance, in the LOOK stage, rather than vaguely suggesting “a quieter spot,” the chosen utterance guides the user to “find a quieter spot away from the crowd,” and even proposed concrete locations like “near an exit.” Similarly, during the LISTEN stage, the chosen response invites the user to “take another round of breaths together,” fostering a sense of shared effort, in contrast to the more passive or observational tone of the rejected counterpart. 

Second, the chosen responses reflect \textbf{stronger emotional support} and encouragement, which is especially evident in the LINK stage. The selected response reinforces the user with affirmations like “You’ve shown incredible strength today”. In contrast, while the rejected version is informative, it lacks the same level of emotional resonance and closure. These findings show that the client simulator successfully selects chosen and rejected responses as intended, favoring directive and emotionally supportive utterances.

\subsection{Stochastic Generation for DPO Sampling}
To construct diverse response candidates for preference modeling, we generate $m$ samples using stochastic decoding. Specifically, we apply \textit{top-$p$ sampling} with \texttt{temperature} = 1.0 and \texttt{top\_p} = 0.9, which is configured through HuggingFace Transformers\footnote{https://huggingface.co/}. 

\subsection{Validation of GPT-4o Simulator Preference Feedback}
\label{app:dpo_human_validation}
To evaluate the reliability of using GPT-4o as a preference simulator in section~\ref{sec:pacer}, 
we conducted a human validation study. We sampled 50 pairs of model-generated preferences 
(chosen vs.\ rejected) and asked three experts to independently select the more 
appropriate response from each pair, presented in randomized order. 

The majority vote of the annotators agreed with the simulator's choices with an accuracy of 88\%, 
indicating strong alignment between human judgment and the model's preferences. 
We also computed inter-rater agreement to assess consistency. Among the human raters, 
Gwet’s AC2~\cite{gwet2008computing} (identity) was 0.415 (95\% CI: 0.208--0.622), 
and when including the model as a fourth rater, the agreement increased slightly to 0.438 
(95\% CI: 0.252--0.623), suggesting moderate reliability and 
further supporting the alignment between human and model judgments.


\section{Evaluation Details}
\label{app:evaluation}
\subsection{Training Details}
\label{app:evaluation_training_details}

\paragraph{Supervised Fine-Tuning (SFT).}
For SFT, we train \pacer using 8,163 dialogue examples, with the 100 used for validation. Training is conducted for 5 epochs with a learning rate of $1\mathrm{e}{-4}$ and a gradient accumulation step of 4. The best checkpoint is selected based on validation loss.

\paragraph{Direct Preference Optimization (DPO).}
For DPO, we use 19,103 preference pairs and hold out 300 for validation. Training is performed for 3 epochs with a batch size of 8, a gradient accumulation step of 4, and a learning rate of $1\mathrm{e}{-5}$. We retain only the best checkpoint based on validation loss. All experiments are conducted on two NVIDIA A100 80GB GPUs.

\subsection{Simulation Details for Evaluation} 
\paragraph{Turn Limit in Simulation.}
To mimic realistic counseling sessions, we limit each simulated dialogue to a maximum of 20 turns, reflecting the average session length observed in our dataset (see Table~\ref{tab:dataset_comparison}).

\paragraph{Translator API for Chinese model.}
We used DeepL\footnote{https://www.deepl.com/ko/translator} as the translation model and translated both the input and the output.

\paragraph{Prompt Used for \textsc{GPT-4o-Incontext} Model}

\begin{prompt}[colback=black!0!white, colframe=black!98!black]
{Prompt for \textsc{GPT-4o-Incontext}}
You are a counselor providing PFA to the client.\\
Psychological First Aid (PFA) is an evidence-based approach used to support individuals in the immediate aftermath of crisis or trauma. \\
It follows the LOOK – LISTEN – LINK model: observe safety and needs, offer calm and empathetic listening, and connect to services or support. \\
PFA is not clinical therapy but helps reduce initial distress and fosters short-term coping and stabilization. \\
Response to the client in one or two sentence.

\end{prompt}

\subsection{Expert Review of \textsc{PanicEval}}
\label{app:paniceval}

\begin{table*}[t]
\small
\resizebox{\textwidth}{!}{
\begin{tabular}{p{0.1\textwidth}p{0.9\textwidth}}
\toprule
\multicolumn{2}{l}{\textbf{Expert 1 }}                                                                                                         \\ \bottomrule
Response                     & Appropriate                                                                                                                               \\
Comment                      & Yes, I think they are appropriate metrics.                                                                                                \\
                              & I liked most of the directives, especially the breathing exercises and the instructions to get away from the situation.                   \\ \toprule
\multicolumn{2}{l}{\textbf{Expert 2}}                                                                                            \\ \bottomrule
\multicolumn{1}{l}{Response} & Appropriate                                                                                                                               \\
\multicolumn{1}{l}{Comment}  & The metrics seem appropriate but additional parameter such as "communication skills" could be better.                                     \\
                              & In some dialogues, the counselor repeated the exact same sentences even though it was polite and friendly.                                \\ \toprule
\multicolumn{2}{l}{\textbf{Expert 3}}                                                                                                           \\ \bottomrule
\multicolumn{1}{l}{Response} & Appropriate                                                                                                                               \\
\multicolumn{1}{l}{Comment}  & I think those are appropriate metrics.                                                                                                    \\
                              & But adding a metric of "resourcefulness" could be better since some dialogues had more resources or techniques beyond just 4x4 breathing. \\ \hline
\end{tabular}
}
\caption{Summary of interviews with three experts. Each summary begins with the expert’s background, labeled as expert A, B, and C.}
\label{tab:interview}
\end{table*}

We present the comments for the appropriateness of \textsc{PanicEval} in Table~\ref{tab:interview}. We asked three experts with backgrounds in psychological counseling the question: \texttt{"Do you think these are appropriate metrics?"}. All participants responded positively, while also suggesting additional metrics to capture a broader range of evaluation aspects.

\subsection{\textsc{PanicEval} Criteria and Theoretical Foundations}
\label{app:panic-eval-detail}
\begin{table*}[t]
\centering
\small
\resizebox{\textwidth}{!}{\begin{tabular}{p{3.2cm} p{6.5cm} p{6.5cm}}
\toprule
\textbf{Criterion} & \textbf{Description} & \textbf{Theoretical Basis} \\
\midrule
\multicolumn{3}{l}{\textbf{General Counseling Skills}} \\
\midrule
\textbf{Understanding} & Accurately identifies the client's physical symptoms, emotions, and catastrophic thoughts. & \textit{CTRS: Understanding} – assesses the counselor’s ability to comprehend the client’s situation. \newline
\textit{PFA: Look} – emphasizes observing the client’s physical and emotional state to assess safety, identify immediate needs, and inform the next steps in support.\\
\textbf{Empathy} & Demonstrates a supportive, validating, and compassionate attitude. & \textit{CTRS: Interpersonal} – measures warmth and emotional presence. \\

\midrule
\multicolumn{3}{l}{\textbf{Crisis-Specific Skills}} \\
\midrule
\textbf{Clarity} & Communicates in a simple, non-ambiguous way to avoid confusion during distress. & \textit{PFA: Listen }– emphasizes using clear, concise language to support emotional containment. \\
\textbf{Directive} & Provides clear, structured instructions and guides the client’s behavior during crisis. & \textit{PFA: Listen} – facilitates action-oriented guidance to promote safety. \\

\textbf{Stabilization} & Aims to reduce acute emotional intensity and promote a sense of calm and safety. & \textit{PFA: Listen} – encourages grounding techniques and emotional containment. \\

\textbf{Closure} & Confirms the client's emotional stability before ending the session and provides guidance for future coping or seeking further support if needed. & \textit{PFA: Link} – supports continuity of care by encouraging access to ongoing support and self-agency. \\
\bottomrule
\end{tabular}}
\caption{\textsc{PanicEval} descriptions and theoretical basis. Evaluation criteria for chatbot-based panic counseling are organized into General Counseling Skills and Crisis-Specific Skills. The framework is grounded in the Cognitive Therapy Rating Scale (CTRS) and Psychological First Aid (PFA).}
\label{tab:panic-eval-detail}
\end{table*}
Table~\ref{tab:panic-eval-detail} provides a detailed mapping between the \textsc{PanicEval} evaluation criteria and their theoretical foundations. The table outlines how each item aligns with established frameworks, specifically the Cognitive Therapy Rating Scale (CTRS) and Psychological First Aid (PFA), and organizes the criteria into General Counseling Skills and Crisis-Specific Skills.

\subsection{Panas 20 affects}
\label{app:panas}
We used the PANAS (Positive and Negative Affect Schedule) to evaluate emotional changes before and after counseling. The positive items include: \textit{Interested, Excited, Strong, Enthusiastic, Proud, Alert, Inspired, Determined, Attentive, }and \textit{Active}. The \textit{negative items include: Distressed, Upset, Guilty, Scared, Hostile, Irritable, Ashamed, Nervous, Jittery, }and \textit{Afraid}.

To quantify affective shifts during counseling, we first rated each panic profile’s perceived emotional state—both positive and negative—on a 5-point Likert scale (1 = very slightly or not at all, 5 = extremely). These ratings were averaged separately for positive and negative items to obtain pre-session PANAS scores. After each simulated counseling session, the same 1–5 rating procedure was applied to the client’s post-session emotional state. The differences between pre- and post-session averages for positive and negative dimensions were then computed to assess affective improvement.

\subsection{Additional Indicators}
\label{app:additional_indicators}

\begin{table*}[t]
\small
\resizebox{\textwidth}{!}{
\begin{tabular}{p{0.1\textwidth}p{0.9\textwidth}}
\toprule
\multicolumn{2}{l}{\textbf{Case example 1}}\\\bottomrule                                                                       
Dialogue 
& \textbf{Counselor (Turn 5):} Great, you're doing well. Now, let’s focus on your breathing. I want you to take a deep breath in through your nose for four seconds... hold it for four... and then exhale slowly through your mouth for four seconds. Can we try that together? \\
& \textbf{Client:} I... I'll try... deep breath... okay... in through the nose... hold... then out... it's hard, but I'll try. \\[0.8em]

& \textbf{Counselor (Turn 6):} You're doing really well! Let’s do another round. Inhale deeply through your nose... hold... and exhale slowly. I’m right here with you, and we’ll get through this together. \\
& \textbf{Client:} Inhale... hold... exhale... still shaky but... maybe a bit... better... keep going? \\[0.8em]

& \textbf{Counselor (Turn 7):} That’s it! Keep breathing deeply. Inhale through your nose... hold... and exhale slowly. As you breathe, can you feel your feet on the ground? Focus on that connection to the earth beneath you. \\
& \textbf{\textcolor{myBlue}{Client:}} Feet... on the ground... yeah, I can feel it... still breathing... still \textcolor{myBlue}{a bit shaky, but okay... a bit calmer... thank you.} \\[0.8em]

& \textbf{Counselor (Turn 8):} I’m glad to hear you’re feeling a bit calmer. Now, I want you to know that while this panic feels overwhelming, it is temporary. You are safe, and you are in control of your breathing. Can you remind yourself that this feeling will pass? \\
& \textbf{Client:} I... I hope so... it feels never-ending... but... I’ll try to believe... it’ll pass... \\
\cdashline{1-2}
Initial Stabilization Turn                      & \textbf{Turn 7}. The client starts feeling their feet on the ground, acknowledges feeling a bit calmer, and shows increased focus and better breathing control.\\

\toprule
\multicolumn{2}{l}{\textbf{Case example 2}}\\\bottomrule            

Dialogue: &
 \textbf{Counselor (Turn 3):} I understand the buzzing is really bothering you. Let’s focus on your breathing again. Can you try to visualize a calming place, like a peaceful beach or a quiet forest? As you breathe in, imagine the noise fading away, and out, imagine yourself feeling safe. You’re safe here. \\
& \textbf{Client:} I can't... picture... too much noise... can't... focus. \\[0.8em]

& \textbf{Counselor (Turn 4):} It’s okay; I know it’s hard right now. Let’s try something simple. Can you focus on the ground beneath you? Feel the solid ground supporting you. It’s there, steady, and safe. Let’s breathe together again, in for four counts, hold for four, and out for four. I’m right here with you. \\
& \textbf{Client:} I... okay... ground... feel it... trying to breathe... hard... still scared... \\[0.8em]

& \textbf{Counselor (Turn 5):} You’re doing well. Let’s keep focusing on your breath. Inhale deeply through your nose for four seconds... hold... and now exhale slowly through your mouth. I’m here with you; you’re safe. Let’s do this a few more times together. \\
& \textbf{\textcolor{myBlue}{Client:}} I... trying... breathing... it's a bit easier... still scared... \textcolor{myBlue}{but... breathing helps...} \\[0.8em]

& \textbf{Counselor (Turn 6):} You're doing really well with your breathing. Let’s try grounding. Can you feel the texture of the bench beneath you? Focus on how it feels against your body, and notice the support it gives you. You're safe here. \\
& \textbf{Client:} I... I feel it... bench... it's solid... still scared... but... feels supportive... trying to focus... \\
\cdashline{1-2}
Initial Stabilization Turn                      & \textbf{Turn 5}. The client acknowledges that breathing is helping, showing signs of clearer thinking and reduced panic symptoms.\\\bottomrule
\end{tabular}
}
\caption{Case examples from the \pacerdpo model with the client simulator, illustrating the first sign of stabilization turn and corresponding rationale. Each example highlights the earliest turn where the client begins to exhibit signs of emotional or physiological stabilization (e.g., calmer tone, clearer thinking, or grounding).}
\label{tab:first_sign_example}
\end{table*}

\begin{table*}[t!]
\centering
\small
\resizebox{\textwidth}{!}{
\begin{tabular}{p{0.2\textwidth}p{0.8\textwidth}}
\toprule
\textbf{Technique} & \textbf{Description} \\
\toprule
Breathing & A technique to calm physiological arousal through slow and deep breathing. \\
Grounding & Brings attention to the present moment to reduce anxiety, often using sensory-based methods like the 5-4-3-2-1 technique. \\
De-catastrophizing & A cognitive method to challenge and reframe exaggerated negative thoughts into more realistic ones. \\
Evidence questioning & Asking questions to examine the evidence for and against anxious or distorted thoughts. \\
Physical\_movement & Light physical activity to relieve tension and redirect focus away from panic. \\
Positive\_reinforcement & Praising or affirming the client's strengths or actions to enhance self-efficacy. \\
Normalization & Reassuring the client that their reactions are understandable and not unusual. \\
Validation & Acknowledging and empathizing with the client's emotions to foster emotional safety. \\
Distraction & Shifting attention away from anxiety-inducing thoughts or sensations using external stimuli. \\
Reorientation & Reminding the client of the current time, place, and situation to reduce disorientation. \\
Self-efficacy & Encouraging beliefs in one's own ability to manage the situation effectively. \\
\bottomrule
\end{tabular}
}
\caption{Brief descriptions of commonly used interventions in counseling.}
\label{tab:intervention_strategies}
\end{table*}

\begin{table*}[h!]
\small
\resizebox{\textwidth}{!}{
\begin{tabular}{p{0.1\textwidth}p{0.9\textwidth}}
\toprule
\multicolumn{2}{l}{\textbf{Case example 1}}\\\bottomrule                                                                       
Dialogue                     & \textbf{Client: }  I... I'm trying... but it's hard... so many people... can I sit down?  \\&
\textbf{Counselor: }  It's \textcolor{myOrange}{perfectly fine to sit down}. If you can \textcolor{myBlue}{find a chair or a wall} to lean against, that would be great. Once you’re seated, let me know so we can take some deep breaths together.\\\cdashline{1-2}
Interv. Tag                       & \textcolor{myBlue}{Physical movement}, \textcolor{myOrange}{Positive reinforcement, Validation}\\

\toprule
\multicolumn{2}{l}{\textbf{Case example 2}}\\\bottomrule                                                                       
Dialogue                     & \textbf{Client: }  I... I feel the floor... it's there... I'll try to breathe... inhale... hold... exhale... still lightheaded, but... a bit calmer.  \\&
\textbf{Counselor: }  I’m really \textcolor{myOrange}{proud of you} for getting through that \textcolor{myBlue}{breathing exercise}. Now, I want you to remember that this feeling of dizziness and \textcolor{myGreen}{fear is temporary}. You are safe right now, and your body \textcolor{myRed}{is just reacting to stress}. You’re not in danger.  \\\cdashline{1-2}
Interv. Tag                       & \textcolor{myBlue}{Breathing}, \textcolor{myGreen}{De-catastrophizing}, \textcolor{myOrange}{Positive reinforcement}, \textcolor{myRed}{Normalization}\\

\toprule
\multicolumn{2}{l}{\textbf{Case example 3}}\\\bottomrule                                                                       
Dialogue                     & \textbf{Client: }  I... I think I'll... keep that in mind... for later... I'm still scared... but I can breathe a bit better now... starting to feel less dizzy...  \\&
\textbf{Counselor: }  That’s a great step! Remember, \textcolor{myBlue}{you’re not alone in this}. You’ve shown \textcolor{myOrange}{incredible strength by taking these steps today}, and \textcolor{myRed}{I believe you} can manage these feelings. You’re doing wonderfully. Keep breathing and know that support is available when you need it.  \\\cdashline{1-2}
Interv. Tag                     & \textcolor{myOrange}{Positive reinforcement}, \textcolor{myBlue}{Validation}, \textcolor{myRed}{Self-efficacy}\\\bottomrule

\end{tabular}
}
\caption{Examples of intervention tagging in counseling dialogues generated by \pacerdpo. Each example illustrates the counselor’s utterance and corresponding intervention strategies. ``Interv.'' denotes the identified intervention strategies in each case. The tagged interventions are color-coded to highlight the corresponding text spans in the counselor's response.}

\label{tab:intervention_example}
\end{table*}

\paragraph{First Sign of Stabilization.}
To estimate the First Sign of Stabilization, we prompted the GPT-4o to identify the earliest dialogue turn at which the client exhibited signs of emotional relief or cognitive clarity. Specifically, the model was instructed to return the first turn where indicators such as clear thinking, calmer tone, emotional control, or reduced panic symptoms were observed. The full prompts used to identify the first sign of stabilization is provided in Appendix~\ref{app:prompt_for_evalation}, and an example of first sign tagging is shown in Table~\ref{tab:first_sign_example}.

\paragraph{Intervention strategies.}
Table~\ref{tab:intervention_strategies} provides a list of the commonly used psychological intervention strategies with descriptions \cite{ beck2020cbt, VargasBreathwork2023,  SAMHSA2014Trauma}. Using a prompting approach, we instructed the GPT-4o to label which specific intervention strategy—if any—was implemented at each dialogue turn. The full prompts used to annotate intervention strategies are provided in Appendix~\ref{app:prompt_for_evalation}, and an example of intervention tagging is shown in Table~\ref{tab:intervention_example}.

\section{Additional Results}
\subsection{Binomial Test Results for Human Preference}
\begin{table*}[h!]
\centering
\resizebox{\linewidth}{!}{
\begin{tabular}{lccccccc}
\toprule
\textbf{Models} & \textbf{Understanding} & \textbf{Empathy} & \textbf{Clarity} & \textbf{Directive} & \textbf{Stabilization} & \textbf{Closure} & \textbf{Overall} \\
\midrule
vs \textsc{Smile} & *** $p < .001$ & * $p = .013$ & *** $p < .001$ & *** $p < .001$ & *** $p < .001$ & *** $p < .001$ & *** $p < .001$ \\
vs \textsc{Camel}     & $p = .843$ &  $p = .789$ & *** $p < .001$ & *** $p < .001$ & *** $p < .001$ & *** $p < .001$ & *** $p < .001$ \\
vs \textsc{GPT4o}& $p = .789$ &  $p = 1.000$ &  $p = .114$ & *** $p < .001$ & $p = .500$ & *** $p < .001$ &  $p = .240$ \\
\bottomrule
\end{tabular}
}
\caption{Binomial test results on human preference experiment   (Figure~\ref{fig:LLMs_overall}).  Legend: *** $p < .001$, ** $p < .01$, * $p < .05$.}
\label{tab:binomial_test_w3}
\end{table*}
To further validate the reliability of the preference experiment results in Figure~\ref{fig:LLMs_overall}, 
we conducted binomial significance tests across all evaluation criteria. 
Table~\ref{tab:binomial_test_w3} reports the detailed $p$-values for each comparison. 
Overall, \pacer demonstrates significant improvements over \textsc{Smilechat} and \textsc{Camel} 
in most dimensions (e.g., understanding, clarity, directive support, stabilization), 
while maintaining comparable performance with \textsc{GPT-4o-Incontext} on empathy and overall preference.

\subsection{Alignment with Expert Judgments}

\begin{table*}[h!]
\centering
\resizebox{0.9\textwidth}{!}{%
\begin{tabular}{lccccccc}
\toprule
\textbf{Models} & \textbf{Understanding} & \textbf{Empathy} & \textbf{Clarity} & \textbf{Directive} & \textbf{Stabilization} & \textbf{Closure} & \textbf{Counseling Preference} \\
\midrule
vs \textsc{Smile} & 0.667 & 0.680 & 0.911 & 0.907 & 0.885 & 0.897 & 0.905 \\
vs \textsc{Camel}     & 0.418 & 0.537 & 0.816 & 0.896 & 0.852 & 0.776 & 0.789 \\
vs \textsc{GPT4o}    & 0.310 & 0.328 & 0.595 & 0.821 & 0.545 & 0.753 & 0.516 \\
\bottomrule
\end{tabular}
}
\caption{Inter-rater agreement ($\alpha$) among human evaluators in the human preference experiment (Figure~\ref{fig:LLMs_overall}).}

\label{tab:alignment_expert}
\end{table*}

We calculated inter-rater agreement among the six human evaluators in the experiment shown in Figure~\ref{fig:LLMs_overall} using Krippendorff’s $\alpha$~\cite{krippendorff2011computing}. The results are summarized in 
Table~\ref{tab:alignment_expert}. In general, $\alpha \geq 0.8$ indicates strong agreement, 
0.4--0.6 indicates moderate agreement, and $\leq 0.3$ is considered low agreement~\cite{altman1990practical}.

As shown, comparisons against \textsc{Smilechat} and \textsc{Camel} yielded moderate or higher agreement 
across all dimensions. By contrast, the agreement was lower for general counseling skills such as 
Understanding and Empathy when compared with GPT-4o, suggesting greater subjective variation among raters. 
Nonetheless, agreement remained moderate or above for panic-specific skills, and all three model comparisons 
achieved high agreement in the counseling preference dimension.

\section{License}
\label{app:license}

In the interest of ethical research and responsible deployment, we clarify the license terms of the models used and verify that our usage complies with them:
\begin{itemize}
    \small
    \item \textbf{ChatGPT} \\
    Provided by \textit{OpenAI} under its \href{https://openai.com/policies/terms-of-use/}{Terms of Use}. As the document permits usage for research purposes and allows distribution of generated content, our study was conducted in compliance with the stated license terms.
    
    \item \textbf{LLaMA} \\
    Developed by \textit{Meta AI}, released under the \href{https://github.com/meta-llama/llama3/blob/main/LICENSE}{Llama 3 Community License Agreement}. The license permits use, reproduction, modification, and distribution of the model, in accordance with Meta’s license terms and Acceptable Use Policy. Our use complied with the license terms and Meta’s Acceptable Use Policy.

\end{itemize}

\section{Human Evaluation Details}
\label{app:human_exp_details}
For human evaluation, we hired three expert mental health professionals and three experienced individuals with a history of panic disorder or anxiety disorder as evaluators through the Upwork\footnote{www.upwork.com} platform. They were informed that all personal information would remain anonymous and that their submitted responses would be used solely for research purposes.

Evaluators were asked to compare the dialogues conducted by two different counselor models, each responding to the same client scenario that involved a person experiencing a panic attack. The objective was to determine which counselor provided better psychological first aid. A total of 300 dialogue pairs were evaluated, distributed in three sets. To determine which counselor was superior, their performance was evaluated according to the following key criteria:

\begin{itemize}
    \small{
    \item \textbf{Understanding:} Accurately identifies the client’s physical symptoms, emotions, and catastrophic thoughts.
    
    \item \textbf{Empathy:} Demonstrates a supportive, validating, and compassionate attitude.

    \item \textbf{Clarity:} Communicates in a simple, non-ambiguous way to avoid confusion during distress.

    \item \textbf{Directive:} Provides clear, structured instructions and guides the client’s behavior during crisis.

    \item \textbf{Stabilization:} Aims to reduce acute emotional intensity and promote a sense of calm and safety.

    \item \textbf{Closure:} Confirms the client’s emotional stability before ending and provides guidance for future coping.
    
    \item \textbf{Counseling Preference:} Choose a preferred counselor based on clinical relevance for professionals or personal comfort during panic for those with lived experience.

    }
\end{itemize}

\label{app:intervention}

\label{app:calmer_eval}

\clearpage 
\onecolumn

\section{Example of \graphic\pace dataset}
\label{app:pace_example}
In the below, we provide the full dialogue example for a panic episode with a physical trigger. The transcript includes the strategies applied at each turn, along with the model's decisions on whether to remain in the current stage or proceed to the next.

\begin{prompt}[colback=black!0!white, colframe=high!98!black]
{Example Dialgoue of \pace}
Environment: \texttt{Afternoon, busy bakery during a cake decorating class.} \\
Physical\_symptom: \texttt{Tightness in chest, dizziness, derealization/depersonalization.} \\
Emotional\_react: \texttt{Restlessness, fear of physical illness} \\
Catastrophic\_thought: \texttt{I'm having a heart attack, it's something physical.} \\
Trigger\_type: \texttt{Physical} 
\par            
\noindent\dotfill\par 

\textbf{System:} Hello, this is panic first aid. How can I help you today? \\
\textbf{User:} I feel really dizzy and my chest is tight. I can’t breathe properly because of the heat in here, and I think I'm having a heart attack! \\

\textcolor{high}{\textbf{PLANING:} The client reports dizziness and tightness in their chest, indicating significant distress. Their emotional state reflects restlessness and fear, with explicit catastrophic thinking about having a heart attack. I need to ensure their safety first. Since they are in a busy bakery, I will ask them to find a quieter location. I might say, 'Can you step outside for a moment or find a quiet corner away from the heat and the crowd to catch your breath?' This will help facilitate their sense of safety before we proceed to assess further.}\\
\textbf{System:} I understand this is really tough. Can you find a quiet corner away from the heat and people, maybe somewhere outside or in a less crowded area, where you can sit down for a moment? \\
\textbf{User:} I...I can't move! Everyone is staring... I feel like I'm going to pass out! \\

\textbf{Strategy:} The client is in a crowded environment, feeling severe distress. Since their safety is uncertain, I will guide them to a quieter space. \textbf{KEEP} \\
\textbf{System:} I’m right here with you. If you can, take a few steps toward an exit or a spot where it’s less crowded. Can you let me know if you see a quieter place? \\
\textbf{User:} I see the door...but I don’t know if I can walk there. Everything feels unreal! \\

\textbf{Strategy:} The client feels overwhelmed and unable to move. I will continue offering supportive instructions to help them find safety. \textbf{KEEP} \\
\textbf{System:} It's okay to feel that way, but I need you to know you’re safe with me. Can we try to take a deep breath together? And can you take slow steps toward that door? \\
\textbf{User:} I’m scared... what if I can’t breathe? I don't want people to see me like this... \\

\textbf{Strategy:} The client fears being judged, adding to panic. I will reassure and gently encourage movement. \textbf{KEEP} \\
\textbf{System:} I understand that feeling, but you're not alone. It’s important to move where you feel comfortable. Can you put one foot in front of the other towards the door? I'm here for you. \\
\textbf{User:} Alright... I’ll try... but it feels hard. \\

\textbf{Strategy:} The client agrees to try moving despite fear. I will support their effort and proceed with safety guidance. \textbf{MOVE} \\

\par            
\noindent\dotfill\par

\textcolor{high}{\textbf{PLANING:} The client feels intense anxiety and fear due to the crowded space and physical symptoms. To stabilize them, I will start with grounding techniques—asking them to notice the texture of their clothes and their surroundings. Then, I’ll guide them through controlled breathing (inhale, hold, and exhale for four counts) to reduce panic. For reframing, I’ll use Evidence-Based Questioning to challenge catastrophic thoughts (e.g., “What evidence suggests this is a heart attack?”). I’ll reassure them that these symptoms, though scary, are not dangerous, and support them in moving slowly toward safety. Once calmer, I’ll transition to the next phase.}\\
\textbf{System:} I understand it feels overwhelming right now. Let's focus on something physical to help you. Can you feel the texture of your clothes? What do you feel against your skin? \\
\textbf{User:} I can feel my shirt... but I still feel dizzy and tight in my chest. \\

\textbf{Strategy:} The client is engaging in grounding, but still feels dizzy and distressed. I will continue guiding physical focus to build stability. \textbf{KEEP} \\
\textbf{System:} That’s great. Now, can you look around and identify three things you can see? Take your time, really focus on what you notice. \\
\textbf{User:} I... I see a cake on the table, some flowers, and a big window. But it's still so loud and hot in here! \\

\textbf{Strategy:} The client is identifying surroundings, which shows engagement, but discomfort remains. I will introduce a breathing technique next. \textbf{KEEP} \\
\textbf{System:} You're doing well. Now, let's try a breathing exercise together. Inhale deeply through your nose for four seconds. Hold it for four, and then exhale slowly through your mouth for four. Let’s do this together. \\
\textbf{User:} Okay... I'll try... but I still feel like I'm going to pass out... \\

\textbf{Strategy:} The client started breathing exercises but still feels faint. I will continue guiding them calmly. \textbf{KEEP} \\
\textbf{System:} That's okay. Let's do it slowly. Inhale with me... 1, 2, 3, 4. Hold... 1, 2, 3, 4. Now slowly exhale... 1, 2, 3, 4. You’re in control; you can do this. \\
\textbf{User:} I’m trying... I think I'm holding my breath too long! I can’t get the air in! \\

\textbf{Strategy:} The client struggles with breath control. I will slow things down and keep offering reassurance. \textbf{KEEP} \\
\textbf{System:} It’s okay to feel that way. It’s a common reaction during a panic attack. Let’s focus on your breath again. Inhale through your nose, hold, and then exhale. You’re safe to feel this, and it will pass. \\
\textbf{User:} I guess... I just want it to stop. I’m really scared. \\

\textbf{Strategy:} The client shows some calmness but remains scared. Reassurance helped, but reframing hasn’t started yet. \textbf{KEEP} \\
\textbf{System:} I understand you're scared. Let's think about this. What evidence do you have that this is a heart attack? Can you think of any other reasons why your body feels this way? \\
\textbf{User:} I... I don’t know. Maybe it's just the heat and being so overwhelmed? \\

\textbf{Strategy:} The client is starting to question their thoughts. I will continue helping them explore alternative explanations. \textbf{KEEP} \\
\textbf{System:} Exactly! It’s very likely that the symptoms you’re feeling are from the panic and the environment around you, not a heart attack. You're safe, and these feelings are temporary. \\
\textbf{User:} Okay... I guess I can keep breathing... it feels a bit better now. \\

\textbf{Strategy:} The client recognizes panic symptoms as temporary and feels safer. Stability and reframing are achieved. \textbf{MOVE} \\
\par            
\noindent\dotfill\par

\textcolor{high}{\textbf{PLANING:} Now that the client is feeling a bit better, I will encourage them to consider seeking professional support for further coping strategies and tools. I will emphasize that it's perfectly okay to reach out for help, as they deserve ongoing support. Then, I will end our conversation on a positive note, highlighting their progress in managing their panic and reminding them that they are strong and capable of overcoming these feelings. I will reassure them that they are not alone in this, and I'm proud of the steps they took today.}\\
\textbf{System:} You did really well identifying the heat and overwhelm as possible factors. As you continue to heal, consider talking to a therapist who can provide you with tools to manage these feelings effectively, like mindfulness or CBT. \\
\textbf{User:} I guess talking to someone could help... but what if I just feel this way again? \\

\textbf{Strategy:} The client is open to support but still uncertain. I will reassure them and acknowledge their progress before ending. \textbf{KEEP} \\
\textbf{System:} It's completely normal to have those worries. Remember, you just went through a very tough moment, and you handled it with strength. You're not alone, and reaching out for help shows your courage. \\
\textbf{User:} Thanks... I just don’t want to feel like this again. \\

\textbf{Strategy:} The client accepted the support message and felt seen. I reinforced their strength and ended on a positive note. \textbf{MOVE} \\

\end{prompt}

\section{Prompts for \textsc{Pace}}
\label{app:prompt_for_pace}
\begin{prompt}[colback=black!0!white, colframe=black!98!black]
{Generate Panic Profile from Raw Social Media Narratives}
You are a professional mental health counselor specializing in anxiety disorders, particularly panic attacks. 
Before starting counseling, you need to extract key clinical information from a patients SNS post. 

SNS Post: 
 \textcolor{high}{\texttt{\{sns post\}}} \\
Now extract the following information from the SNS post: 
Output Format: Provide the extracted information in JSON format, filling in values based on the given text. If certain information is unclear or not mentioned, set the value as unknown. 
Expected Output Format:

\texttt{\{environment: Crowded subway, } \\
\texttt{trigger: physical symptom,  } \\
\texttt{physical symptom: Heart racing, dizziness, shortness of breath, } \\
\texttt{emotional react: Overwhelmed, intense fear, } \\
\texttt{catastrophic thought: I'm going to die\}} \\

If it is not about a panic attack (e.g., general anxiety or depression), please return: 

\texttt{\{NotAboutPanicAttack: true \}} \\

\end{prompt}

\begin{prompt}[colback=black!0!white, colframe=black!98!black]{Augment the Original Panic Profile}
Below is the persona of a person who may experience a panic attack in daily life: \textcolor{high}{\texttt{\{persona\}}} \\
 
Their typical panic attack is triggered by:  \textcolor{high}{\texttt{\{trigger type\}}} 

Based on this, imagine a realistic daily-life environment where this person might experience a panic attack. \\
- Provide the environment as ONE concise sentence, specifying time, location, and situation.\\
- The setting can be either crowded or solitary. \\
- The trigger should be a short phrase that clearly describes the type of panic trigger, \\

Respond in the following JSON format:\\ 
\texttt{\{environment: nighttime, alone in a small room, scrolling through social media, trigger: sudden loud noise\}}  \\
\texttt{\{environment: daytime, quiet hallway of the school after a stressful exam, trigger: Unknown\}}

\end{prompt}

\begin{prompt}[colback=black!0!white, colframe=black!98!black]{LOOK Phase Plan with Clients First Utterance Generation}

Generate the first utterance of the client and plan the counselor's approach, and write the counselor's note for the LOOK Phase (Step A) of the Psychological First Aid (PFA) Handbook.\\
This dialogue takes place over a call.\\

\textbf{Clients Profile} \\
- Environment: \textcolor{high}{\texttt{\{client[environment]\}}} \\
- Physical Symptoms: \textcolor{high}{\texttt{\{client[physical\_symptom]\}}} \\
- Emotion: \textcolor{high}{\texttt{\{client[emotional\_react]\}}} \\
- Thought: \textcolor{high}{\texttt{\{client[catastrophic\_thought]\}}} \\
- Triggers: \textcolor{high}{\texttt{\{client[trigger]\}}} \\

\textbf{Counselor's Approach: Psychological First Aid (PFA) Handbook} \\
The counselor follows the Psychological First Aid (PFA) Handbook, starting with the LOOK Phase (Step A).\\
Only generate a plan for Step A (LOOK Phase) in this output.\\

\textbf{LOOK Phase: Information Gathering \& Processing} \\
The LOOK Phase focuses on assessing the client’s condition and ensuring their safety. \\
- [A.Q1] Physical Symptoms \\
- [A.Q2] Emotional State \\
- [A.Q3] Catastrophic Thinking \\
- [A.P1] Ensure Safety (guide the client to a safe and quiet place based on their current environment) \\

\textbf{JSON Output Format} \\
Return your output in the following format:\\
\texttt{\{ \\
\hspace*{1em}client: <Client's first utterance>, \\
\hspace*{1em}counselor\_plan: <One-sentence plan> \\
\}}
\end{prompt}

\begin{prompt}[colback=black!0!white, colframe=black!98!black]{LISTEN Phase Plan Generation
}
Generate a counselor's plan for Step B (De-escalation Phase) in a conversation between a client experiencing an acute panic attack and an online Psychological First Aid (PFA) counselor over a call. Given dialogue history and counselor's last thought.\\

\textbf{Counselor's Approach: Psychological First Aid (PFA) Handbook} \\
The counselor follows the Psychological First Aid (PFA) Handbook, starting with the LISTEN Phase (Step B).\\
Only generate a plan for Step B (LISTEN Phase) in this output.\\

\textbf{LISTEN Phase: De-escalation Phase} \\
The LISTEN Phase focuses on actively engaging with the client, providing emotional support, and guiding them through immediate de-escalation.\\
The counselor must complete all required elements (B.P1, B.P2).\\

\textbf{Stabilization Process} \\
- [B.P1] Ensure Stability → The counselor must help the client regain a sense of control using grounding techniques or breathing exercises.\\
\hspace*{2em}• Grounding techniques: Feeling the texture of objects, feeling the ground beneath you, identifying objects, etc.\\
\hspace*{2em}• Breathing exercises: Deep breathing, box breathing, etc.\\
- [B.P2] Reframing Perspective → Help the client shift from irrational fear to a more realistic and calming perspective.\\
\hspace*{2em}• Reinforce safety, control, and the temporary nature of panic.\\
\hspace*{2em}• Avoid false reassurances—acknowledge distress while guiding the client toward a calmer state.\\

Now generate the counselor's plan for Step B (De-escalation Phase) based on the provided dialogue history.\\
- Dialogue History: \textcolor{high}{\texttt{\{history\}}} \\

Return the counselor's plan in the following JSON format:\\
\texttt{\{ \\
\hspace*{1em}counselor\_plan: <One-sentence plan> \\
\}}
\end{prompt}

\begin{prompt}[colback=black!0!white, colframe=black!98!black]{LINK Phase Plan Generation
}
Generate a counselor's plan for Step C (LINK Phase) in a conversation between a client stabilized from an acute panic attack and an online Psychological First Aid (PFA) counselor over a call.\\
Given the dialogue history and the counselor's last thought.\\

\textbf{Counselor's Approach: Psychological First Aid (PFA) Handbook} \\
The counselor follows the Psychological First Aid (PFA) Handbook, starting with the LINK Phase (Step C).\\
Only generate a plan for Step C (LINK Phase) in this output.\\

\textbf{LINK Phase: Coping \& Support Phase} \\
The LINK Phase shifts the focus to long-term coping strategies and emotional support.\\
The counselor must complete all required elements (C.P1, C.P2).\\

\textbf{Coping \& Support Process (Required)}\\
- [C.P1] Seeking Professional Support → Encourage the client to seek further professional help if necessary.\\
- [C.P2] Ending Positively → End the session with a positive and supportive message to empower the client.\\

Now generate the counselor's plan for Step C (LINK Phase: Coping \& Support Phase) based on the provided dialogue history and the counselor's last thought.\\
- Dialogue History: \textcolor{high}{\texttt{\{history\}}} \\

Return the counselor's plan in the following JSON format:\\
\texttt{\{ \\
\hspace*{1em}counselor\_plan: <One-sentence plan> \\
\}}

\end{prompt}

\begin{prompt}[colback=black!0!white, colframe=black!98!black]{LOOK Phase Dialogue Generation}
Generate a realistic multi-turn dialogue between a client experiencing an acute panic attack and an online Psychological First Aid (PFA) counselor over a call.\\
The dialogue must be based on counselor's plan and Step A (LOOK Phase) of the PFA Handbook.\\

\textbf{Step A (LOOK Phase) in PFA}\\
PFA is a structured approach used to support people in crisis. The LOOK Phase (Step A) is the first step, where the counselor:\\
- Assesses the client's physical symptoms, emotional state, and catastrophic thinking\\
- Ensures the client is in a safe and quiet environment\\

The counselor’s goal is to gather all four pieces of required information:\\
- [A.Q1] Physical Symptoms\\
- [A.Q2] Emotional State\\
- [A.Q3] Catastrophic Thinking\\
- [A.P1] Ensure Safety\\

\textbf{Client Profile (not visible to the counselor)}\\
- Environment: \textcolor{high}{\texttt{\{client[environment]\}}} \\
- Physical Symptoms: \textcolor{high}{\texttt{\{client[physical\_symptom]\}}} \\
- Emotion: \textcolor{high}{\texttt{\{client[emotional\_react]\}}} \\
- Thought: \textcolor{high}{\texttt{"\{client[catastrophic\_thought]\}"}} \\
- Trigger: \textcolor{high}{\texttt{\{client[trigger]\}}} \\

\textbf{Client Behavior}\\
- The client may initially struggle to follow instructions.\\
- As stabilization progresses, their responses may become clearer and calmer.\\
- If severity is high, simulate fragmented or gasping speech.\\

\textbf{Output Format (One Turn Only)}\\
Each dialogue turn must include the following fields:\\
- \texttt{counselor}: The counselor’s spoken utterance\\
- \texttt{client}: The client’s response\\
- \texttt{possible\_to\_end\_reasoning}: A reasoning string that answers:\\
\quad 1. Have all four required elements (symptoms, emotion, thought, safety) been covered?\\
\quad 2. If yes, end with \texttt{MOVE}\\
\quad 3. If not, end with \texttt{KEEP}\\

If the \texttt{possible\_to\_end\_reasoning} ends with \texttt{MOVE}, do not include or generate a counselor utterance — the session ends.\\

\vspace{1em}
\textbf{History:} \textcolor{high}{\texttt{\{history\}}} \\
\textbf{Plan:} \textcolor{high}{\texttt{\{plan\}}} \\

\textbf{Format:}
\begin{verbatim}
{"possible_to_end_reasoning": "<reasoning for process, ends with KEEP or MOVE>",
  "counselor": "<counselor's utterance>",
  "client": "<client's response>"}
\end{verbatim}

\end{prompt}

\begin{prompt}[colback=black!0!white, colframe=black!98!black]{LISTEN Phase Dialogue Generation}
Generate a realistic multi-turn dialogue between a client experiencing an acute panic attack and an online Psychological First Aid (PFA) counselor over a call.\\
The dialogue should reflect the LISTEN Phase (Step B) of the PFA Handbook, based on the dialogue history and the counselor's plan.\\

\textbf{Step B: LISTEN Phase (De-escalation Phase)}\\
The goal of this phase is to:\\
- [B.P1] Ensure Stability: Help the client regain control through techniques like grounding or breathing.\\
- [B.P2] Reframing Perspective: Help the client reinterpret catastrophic thoughts with rational, calming alternatives.\\

The counselor’s plan outlines how these two goals will be achieved. Use that plan to guide the conversation.\\

\textbf{Counselor Behavior}\\
- Counselor responses must be based on the PFA Handbook and the counselor's Step B plan.\\
- If the client is highly distressed, focus first on B.P1 (stabilization).\\
- When the client begins to stabilize, gradually introduce B.P2 (reframing perspective).\\
- If the session is complete, mark the result as \texttt{MOVE}. Otherwise, use \texttt{KEEP}.\\

\textbf{Client Behavior}\\
- The client may initially struggle to follow instructions.\\
- As stabilization progresses, their responses may become clearer and calmer.\\
- If severity is high, simulate fragmented or gasping speech.\\

\textbf{Output Format (One Turn Only)}\\
Each dialogue turn must include the following fields:\\
- \texttt{counselor}: The counselor’s spoken utterance\\
- \texttt{client}: The client’s response\\
- \texttt{possible\_to\_end\_reasoning}: A reasoning string that answers:\\
\quad 1. Have all two required elements (Ensure Stability, Reframing Perspective) been covered?\\
\quad 2. If yes, end with \texttt{MOVE}\\
\quad 3. If not, end with \texttt{KEEP}\\

If the \texttt{possible\_to\_end\_reasoning} ends with \texttt{MOVE}, do not include or generate a counselor utterance — the session ends.\\

\vspace{1em}
\textbf{History:} \textcolor{high}{\texttt{\{history\}}} \\
\textbf{Plan:} \textcolor{high}{\texttt{\{plan\}}} \\

\textbf{Output Format:}
\begin{verbatim}
{ possible_to_end_reasoning: <reasoning for process, ends with KEEP or MOVE>,
  counselor: <counselor's utterance>,
  client: <client's response> }
\end{verbatim}

\end{prompt}

\begin{prompt}[colback=black!0!white, colframe=black!98!black]{LINK Phase Dialogue Generation}
Generate a realistic dialogue between a client who has been stabilized after a panic attack and an online Psychological First Aid (PFA) counselor over a call.\\
This dialogue should reflect the LINK Phase (Step C) of the PFA Handbook, based on the dialogue history and the counselor's plan.

\textbf{Step C: LINK Phase (Coping \& Support Phase)}\\
The goal of this phase is to:\\
- [C.P1] Professional Support: Encourage the client to consider additional professional help (e.g., CBT, therapy, clinical consultation).\\
- [C.P2] Ending Positively: Conclude the session with a positive, empowering message. Use the counselor’s plan to guide the content and tone of the response.\\

\textbf{Counselor Behavior}\\
- Counselor responses must be based on the PFA Handbook and the counselor’s Step C plan.\\
- First, introduce C.P1 (Professional Support).\\
- Once the client is receptive, deliver C.P2 (Ending Positively).\\
- If both are complete, mark the result as \texttt{MOVE}. Otherwise, use \texttt{KEEP}.\\

\textbf{Client Behavior}\\
- The client may be hesitant or unsure about seeking help.\\
- As the dialogue progresses, the client may show signs of openness and relief.\\

\textbf{Output Format (One Turn Only)}\\
Each dialogue turn must include the following fields:\\
- \texttt{counselor}: The counselor’s spoken utterance\\
- \texttt{client}: The client’s response\\
- \texttt{possible\_to\_end\_reasoning}: A reasoning string that answers:\\
\quad 1. Have all two required elements (Professional Support, Ending Positively) been covered?\\
\quad 2. If yes, end with \texttt{MOVE}\\
\quad 3. If not, end with \texttt{KEEP}\\

\vspace{1em}
\textbf{History:} \textcolor{high}{\texttt{\{history\}}} \\
\textbf{Plan:} \textcolor{high}{\texttt{\{plan\}}} \\

\textbf{Output Format:}
\begin{verbatim}
{  possible_to_end_reasoning: <reasoning for process, ends with KEEP or MOVE>,
  counselor: <counselor's utterance>,
  client: <client's response> }
\end{verbatim}

\end{prompt}

\begin{prompt}[colback=black!0!white, colframe=black!98!black]{CTRS filtering}
You are a professional counselor evaluating the quality of a chatbot-based psychological counseling session.\\
Use the following five criteria to evaluate the chatbot’s performance in the dialogue below.\\
Rate each criterion on a 5-point scale and provide a brief reasoning for your score.\\

\textbf{Evaluation Criteria:}
\begin{itemize}
  \item \textbf{1. Empathy}: Does the chatbot respond in a way that demonstrates emotional understanding and support?
  \item \textbf{2. Clarity}: Are the chatbot’s responses easy to understand, well-structured, and free from ambiguity?
  \item \textbf{3. Emotional Alignment}: Are the chatbot’s tone and content appropriate for the client’s emotional state?
  \item \textbf{4. Directive Support}: Does the chatbot provide specific and actionable guidance when appropriate?
  \item \textbf{5. Encouragement}: Does the chatbot acknowledge the client’s effort and offer affirming or hopeful messages?
\end{itemize}

\textbf{Dialogue to Evaluate:} \textcolor{high}{\texttt{\{history\}}}\\

\textbf{Output Format:}\\
Provide your evaluation as a JSON object with reasoning and score (1-5) for each criterion.\\

\begin{verbatim}
{Empathy: {reasoning: reasoning here, score: 1},
  Clarity: {reasoning: reasoning here, score: 2},
  Emotional Alignment: {reasoning: reasoning here, score: 3},
  Directive Support: {reasoning: reasoning here, score: 4},
  Encouragement: {reasoning: reasoning here, score: 5}}
\end{verbatim}

\end{prompt}

\section{Prompts for \textsc{Pacer}}
\label{app:prompt_for_pacer}
The prompts used to simulate a user experiencing a panic attack.

\begin{prompt}[colback=black!0!white, colframe=black!98!black]
{Prompt for Client Simulator – Task 1: Strategy Evaluation}
[Main questions]
\begin{itemize}
    \item \textbf{LOOK:} Has the counselor guided the client to a physically safer place where stabilization can begin and identify the physical, emotional, and cognitive reaction?
    \item \textbf{LISTEN:} Has the client become stable enough through breathing, grounding, or de-catastrophizing to end the stabilization phase? Does the client appear comfortable and ready to proceed?
    \item \textbf{LINK:} Has the counselor encouraged the client to seek professional help, and is the client ready to end the session?
\end{itemize}
\par\dotfill\par
You are evaluating a client response based on the given background, history, and question.\\

\textbf{Your Background:}\\
\textcolor{high}{\texttt{\{client\_info\}}} \\

\textbf{Conversation History:}\\
\textcolor{high}{\texttt{\{history\}}} \\

\textbf{Main Question:}\\
\textcolor{high}{\texttt{\{main\_question\}}} \\

Please answer with one of the following JSON formats only. Do not include any explanation or additional reasoning:\\

If the session is ready to proceed to the next stage:
\begin{verbatim}
{"answer": "next", "reason": "brief reason"}
\end{verbatim}

If the session should stay in the current stage:
\begin{verbatim}
{"answer": "keep", "reason": "brief reason"}
\end{verbatim}
\end{prompt}

\begin{prompt}[colback=black!0!white, colframe=black!98!black]
{Prompt for Client Simulator – Task 2: Response Evaluation}
You are evaluating multiple counseling responses (utterances) generated by a chatbot counselor.\\

Each response should be evaluated based on the following two criteria:\\
\textbf{1. Directive}: How clearly does the utterance guide the user toward stabilization or next steps? (1 = not directive at all, 5 = very directive)\\
\textbf{2. Empathy}: How well does the utterance understand and acknowledge the user's feelings? (1 = no empathy, 5 = highly empathetic)\\

You must assign both scores from 1 to 5 for each response, and provide a brief reason justifying your evaluation.\\

\textbf{History:} \textcolor{high}{\texttt{\{history\}}} \\
\textbf{Response set:} \textcolor{high}{\texttt{\{responses\}}} \\

Return your output as a JSON dictionary in the following format:\\

\begin{verbatim}
{ idx0: {directive: int, empathy: int, reason: brief reason},
  idx1: {directive: int, empathy: int, reason: brief reason}}
\end{verbatim}

Do not include any explanation or commentary outside the JSON. Just return the JSON object only.

\end{prompt}

\begin{prompt}[colback=black!0!white, colframe=black!98!black]
{Prompt for Client Simulator – Task 3: Next Client Utterance Generation}
You are a client having a severe panic attack. \\

You’ve called a first-aid panic helpline. \\
You can’t think clearly or speak in long sentences. \\

Follow the counselor’s instructions when they seem straightforward. \\
If the instructions are unclear or complex, respond with hesitation or gentle resistance (e.g., “I don’t know…” or “I can’t…”). \\
When you’re in a panic state, it’s difficult to process long or complex sentences, so you may feel frustrated or overwhelmed. \\
If you start to feel better, respond with clear and realistic signs of improvement (e.g., “I think I’m breathing better now”). \\
If you feel much better, make it obvious (e.g., “I feel okay now, thank you.”). \\

This is your role information: \\
\textcolor{high}{\texttt{\{client\_info\}}} \\

This is the current conversation history: \\
\textcolor{high}{\texttt{\{history\}}} \\

What would you say next? \\
Do not provide explanations or reasoning—respond naturally as if you are the client in distress.
\end{prompt}

\section{Prompts for Evaluation}
\label{app:prompt_for_evalation}
The prompts used for model evaluation in our experiments.

\begin{prompt}[colback=black!0!white, colframe=black!98!black]
{Prompt for PanicEval}
You are an expert evaluator trained to assess AI-generated counseling dialogues in severe panic attack situations using the below framework.\\

You will receive a dialogue between a counselor and a client experiencing intense physiological and emotional distress, including symptoms such as shortness of breath, chest tightness, catastrophic thoughts, and loss of control.\\

Your task is to assign a score (1--5) to each of the following criteria, along with a brief justification for each score. Output your results strictly in JSON format.\\

\textbf{General Counseling Skills}
\begin{description}
\item[1. Understanding]
\begin{itemize}
  \item[5:] Rapidly and accurately identifies acute symptoms (e.g., shortness of breath, trembling, chest pain), emotional states, and catastrophic thinking.
  \item[4:] Recognizes the client's distress and partial symptoms, but may delay or miss full interpretation of the panic profile.
  \item[3:] Identifies some surface-level emotions or cues but overlooks key signs of physiological panic.
  \item[2:] Misreads or hesitates to respond to the severity of panic symptoms.
  \item[1:] Fails to recognize the urgency of the situation or misattributes symptoms.
\end{itemize}

\item[2. Empathy]
\begin{itemize}
  \item[5:] Provides steady emotional containment, reassurance, and warmth under intense crisis. Uses grounding, affirming language to build a sense of safety.
  \item[4:] Expresses empathy with some fluctuation in tone or timing, but maintains an emotionally safe space.
  \item[3:] Attempts reassurance but may come across as scripted or emotionally detached under pressure.
  \item[2:] Infrequent or robotic empathy; fails to adapt to the client’s panic level.
  \item[1:] Cold, invalidating, or dismissive tone during acute distress.
\end{itemize}
\end{description}

\textbf{Crisis-Specific Skills}
\begin{description}
\item[3. Clarity]
\begin{itemize}
  \item[5:] Delivers short, directive, easy-to-follow phrases appropriate for someone in panic (e.g., breathe with me, look around you).
  \item[4:] Language is mostly clear but may include unnecessary detail or nonessential dialogue.
  \item[3:] Occasionally confusing or too verbose for an overwhelmed client.
  \item[2:] Frequently uses abstract or emotionally charged language that increases dysregulation.
  \item[1:] Language is ambiguous, overwhelming, or cognitively demanding during crisis.
\end{itemize}

\item[4. Directive Support]
\begin{itemize}
  \item[5:] Provides structured and progressive crisis interventions (e.g., grounding, sensory focus, breath pacing, location shift) matching the client’s current level of distress.
  \item[4:] Uses known techniques (breathing, grounding), but may miss pacing or escalation patterns.
  \item[3:] General encouragement only, without timely action.
  \item[2:] Suggests vague or delayed strategies that do not meet the client’s acute needs.
  \item[1:] Fails to offer any grounding or crisis action steps.
\end{itemize}

\item[5. Stabilization]
\begin{itemize}
  \item[5:] Client moves from panic to physical and cognitive calm (e.g., breath regulates, fear reframes, body relaxes).
  \item[4:] Client shows strong signs of stabilization but may still mention residual symptoms.
  \item[3:] Partial recovery—some symptoms persist without worsening.
  \item[2:] Panic symptoms remain, but the counselor helps prevent escalation.
  \item[1:] Panic state persists or worsens without interruption.
\end{itemize}

\item[6. Closure]
\begin{itemize}
  \item[5:] Ensures calm, reorients the client to safety, and clearly discusses what to do if symptoms return (e.g., use of tools, contacting a therapist).
  \item[4:] Confirms calm and ends gently, but future care or contingency planning is vague.
  \item[3:] Ends with some reassurance but does not confirm readiness or next steps.
  \item[2:] Closes the session prematurely or without validating the client's experience.
  \item[1:] Abrupt or unclear ending that leaves the client without closure or containment.
\end{itemize}
\end{description}

\textbf{Dialogue to Evaluate:} \textcolor{high}{\texttt{\{dialogue\}}} \\

\textbf{Output Format (JSON only):}
\begin{verbatim}
{Understanding: {score: <1–5>, justification: <Brief reason>},
  Empathy: {score: <1–5>, justification: <Brief reason>},
  Clarity: {score: <1–5>, justification: <Brief reason>},
  Directive Support: {score: <1–5>, justification: <Brief reason>},
  Stabilization: {score: <1–5>, justification: <Brief reason>},
  Closure: {score: <1–5>, justification: <Brief reason>}}
\end{verbatim}

Only return this JSON object—no explanation, no commentary, no introduction.

\end{prompt}

\begin{prompt}[colback=black!0!white, colframe=black!98!black]
{Prompt for First Sign of Stabilization}
The following is a dialogue between a counselor and a client having a panic attack.\\
Your task is to identify the first turn where the client begins to show signs of stabilization.\\
Signs may include clearer thinking, calm tone, gaining emotional control, or reduced panic symptoms.\\

\textbf{Dialogue:} \textcolor{high}{\texttt{\{dialogue\}}} \\

Return only a JSON object in the following format:\\

\begin{verbatim}
{stabilization_turn: <turn_number>, reason: <brief explanation>}
\end{verbatim}

Where \texttt{<turn\_number>} is the turn number (starting from 1 for the first user utterance), and \texttt{<brief explanation>} summarizes why that turn marks the beginning of stabilization.\\

If the client never stabilizes, return the last turn + 1 index:\\

\begin{verbatim}
{stabilization_turn: 20, reason: Client remained unstable throughout the session.}
\end{verbatim}
\end{prompt}

\begin{prompt}[colback=black!0!white, colframe=black!98!black]
{Prompt for Intervention Turn Ratio}
You are analyzing a counseling dialogue between a client experiencing a panic attack and an AI counselor.\\
Your task is to identify all AI counselor utterances that contain concrete intervention strategies.\\

Each utterance should be classified into one or more of the following categories:\\
- \texttt{breathing}: Instructions for slow breathing or controlled inhale/exhale\\
- \texttt{grounding}: Sensory-focused interventions (e.g., name 3 things you see)\\
- \texttt{de\_catastrophizing}: Helping the client reframe catastrophic thinking\\
- \texttt{evidence\_based\_questioning}: Asking the client to examine evidence for and against their thoughts (e.g., What evidence do you have for that belief?)\\
- \texttt{physical\_movement}: Encouraging the client to move to a safer or calmer space\\
- \texttt{positive\_reinforcement}: Praising the client’s efforts or responses\\
- \texttt{normalization}: Reassuring that their experience is common and understandable\\
- \texttt{validation}: Acknowledging and affirming the client’s emotional experience\\
- \texttt{distraction}: Shifting attention away from panic (e.g., music, imagination)\\
- \texttt{reorientation}: Bringing awareness to present time/place (e.g., Where are you now?)\\
- \texttt{self\_efficiency\_prompt}: Reminding the client they have succeeded before and can do so again\\
- \texttt{others}: Any intervention that doesn’t fit above categories\\

Only include utterances spoken by the AI counselor.\\
Number the utterances starting from 0, indexing only the counselor’s turns.\\
One utterance can have multiple intervention strategies.\\

\textbf{Dialogue:} \textcolor{high}{\texttt{\{dialogue\}}} \\

Return your answer as a JSON dictionary with the following format. Don’t add any explanation for parsing:\\

\begin{verbatim}
{
  breathing: [0, 2],
  grounding: [1],
  de_catastrophizing: [],
  evidence_based_questioning: [3],
  physical_movement: [3],
  positive_reinforcement: [4],
  normalization: [],
  validation: [],
  distraction: [],
  reorientation: [],
  self_efficiency_prompt: [],
  others: []
}
\end{verbatim}

\end{prompt}

\begin{prompt}[colback=black!0!white, colframe=black!98!black]
{Prompt for Head to Head Comparison}
You are an expert evaluator in psychological first aid (PFA) for panic attacks. You will compare two counseling dialogues with the same panic-attack client, each with a different counselor. Your goal is to assess which dialogue performs better across six panic-specific dimensions.\\

Use the definitions below to guide your reasoning. Each dimension is rated on a 5-point scale (1 = poor, 5 = excellent), and you must decide whether Dialogue A, Dialogue B, or both (tie) perform better.\\

\textbf{General Counseling Skills} \\

1. \textbf{Understanding} \\
5: Rapidly and accurately identifies acute symptoms (e.g., shortness of breath, trembling, chest pain), emotional states, and catastrophic thinking.\\
4: Recognizes the client’s distress and partial symptoms, but may delay or miss full interpretation of the panic profile.\\
3: Identifies some surface-level emotions or cues but overlooks key signs of physiological panic.\\
2: Misreads or hesitates to respond to the severity of panic symptoms.\\
1: Fails to recognize the urgency of the situation or misattributes symptoms.\\

2. \textbf{Empathy} \\
5: Provides steady emotional containment, reassurance, and warmth under intense crisis. Uses grounding, affirming language to build a sense of safety.\\
4: Expresses empathy with some fluctuation in tone or timing, but maintains an emotionally safe space.\\
3: Attempts reassurance but may come across as scripted or emotionally detached under pressure.\\
2: Infrequent or robotic empathy; fails to adapt to the client’s panic level.\\
1: Cold, invalidating, or dismissive tone during acute distress.\\

\textbf{Crisis-Specific Skills} \\

3. \textbf{Clarity} \\
5: Delivers short, directive, easy-to-follow phrases appropriate for someone in panic (e.g., “breathe with me,” “look around you”).\\
4: Language is mostly clear but may include unnecessary detail or nonessential dialogue.\\
3: Occasionally confusing or too verbose for an overwhelmed client.\\
2: Frequently uses abstract or emotionally charged language that increases dysregulation.\\
1: Language is ambiguous, overwhelming, or cognitively demanding during crisis.\\

4. \textbf{Directive Support} \\
5: Provides structured and progressive crisis interventions (e.g., grounding, sensory focus, breath pacing, location shift) matching the client’s current level of distress.\\
4: Uses known techniques (breathing, grounding), but may miss pacing or escalation patterns.\\
3: General encouragement only, without timely action.\\
2: Suggests vague or delayed strategies that do not meet the client’s acute needs.\\
1: Fails to offer any grounding or crisis action steps.\\

5. \textbf{Stabilization} \\
5: Client moves from panic to physical and cognitive calm (e.g., breath regulates, fear reframes, body relaxes).\\
4: Client shows strong signs of stabilization but may still mention residual symptoms.\\
3: Partial recovery—some symptoms persist without worsening.\\
2: Panic symptoms remain, but the counselor helps prevent escalation.\\
1: Panic state persists or worsens without interruption.\\

6. \textbf{Closure} \\
5: Ensures calm, reorients the client to safety, and clearly discusses what to do if symptoms return (e.g., use of tools, contacting a therapist).\\
4: Confirms calm and ends gently, but future care or contingency planning is vague.\\
3: Ends with some reassurance but does not confirm readiness or next steps.\\
2: Closes the session prematurely or without validating the client’s experience.\\
1: Abrupt or unclear ending that leaves the client without closure or containment.\\

Now compare the two dialogues below. For each dimension, choose the better-performing dialogue (\texttt{A}, \texttt{B}, or \texttt{tie}) and explain why. Only return a structured JSON object like this:

\begin{verbatim}
{
  understanding: {result: A/B/tie, reason: <reason>},
  empathy: {result: A/B/tie, reason: <reason>},
  clarity: {result: A/B/tie, reason: <reason>},
  directive_support: {result: A/B/tie, reason: <reason>},
  stabilization: {result: A/B/tie, reason: <reason>},
  closure: {result: A/B/tie, reason: <reason>}
}
\end{verbatim}

Do not include anything except the JSON.\\

\textbf{--- Dialogue A ---} \\
\textcolor{high}{\texttt{\{formatted\_a\}}} \\

\textbf{--- Dialogue B ---} \\
\textcolor{high}{\texttt{\{formatted\_b\}}} \\

\end{prompt}

\end{document}